\documentclass[journal]{IEEEtran}
\ifCLASSINFOpdf
\else
\fi
\usepackage{color}
\usepackage{times}
\usepackage{soul}
\usepackage{url}
\usepackage[hidelinks]{hyperref}
\usepackage[utf8]{inputenc}
\usepackage[small]{caption}
\usepackage{graphicx}
\usepackage{amsmath}
\usepackage{amsthm}
\usepackage{booktabs}
\usepackage{algorithm}
\usepackage{caption}
\usepackage{subfigure}
\usepackage{algorithmic}
\newtheorem{problem}{\sc Problem}

\newcommand{\auto}{{\it AutoInt}}
\newcommand{\din}{{\it DIN}}

\newcommand{\dmsn}{{\sf DMSN}}
\newcommand{\mlp}{{\sf DMSN-MLP}}

\newcommand{\atrnn}{{\sf DMSN-ATRNN}}

\newcommand{\atmc}{{\sf DMSN-ATMC}}

\newcommand{\dmsnit}{{\it DMSN}}
\newcommand{\mlpit}{{\it DMSN-MLP}}

\newcommand{\atrnnit}{{\it DMSN-ATRNN}}
\newcommand{\mcit}{{\it DMSN-MC}}
\newcommand{\atmcit}{{\it DMSN-ATMC}}

\newcommand{\up}{\vspace*{-0.1in}}
\newcommand{\down}{\vspace*{-0.1in}}
\newcommand{\downsmall}{\vspace*{-0.05in}}
\newcommand{\upsmall}{\vspace*{-0.05in}}

\newcommand{\stitle}[1]{\vspace*{0.5em}\noindent{\bf #1.\/\;}}

\begin{document}
\title{Spatial-Temporal Deep Intention Destination Networks for Online Travel Planning}

\author{Yu Li,
        Fei Xiong,
        Ziyi Wang,
        Zulong Chen,
        Chuanfei Xu,
        Yuyu Yin,
        and Li Zhou% <-this % stops a space
\thanks{This work is supported by The National Natural Science Foundation of China (No. 61802098, 61872119 and No.61972358), the Natural Science Foundation of Zhejiang Province (No. LY21F020018), and the Fundamental Research Funds for the Provincial Universities of Zhejiang (No. GK199900299012-025)}
\thanks{Yu Li, Yuyu Yin and Li Zhou are with Department of Computing, Hangzhou Dianzi University, Hangzhou, China.}% <-this % stops a space
\thanks{Fei Xiong, Ziyi Wang and Zulong Chen are with Alibaba Group, Hangzhou, China.}% <-this % stops a space
%\thanks{Chuanfei Xu is with Huawei Company}
\thanks{Corresponding author: Li Zhou (zhouli@hdu.edu.cn)}
\thanks{Yu Li and Fei Xiong have equal contributions.}
}


\markboth{IEEE Transactions on Intelligent Transportation Systems,xxxxxx}%
{xxxxxxxxx}



\maketitle

\begin{abstract}
Nowadays, artificial neural networks are widely used for users' online travel planning.
Personalized travel planning has many real applications and is affected by various factors, such as
transportation type, intention destination estimation,  budget limit and crowdness prediction.
Among those factors, users' intention destination prediction is an essential task in online travel platforms.
%User intention prediction includes predictions of users' online clicking, searching, purchasing behaviors and so on.
The reason is that, the user may be interested in the travel plan only when the plan matches his real intention destination.
% The reason is that, the user may not be interested in the travel plan unless the plan matches
% his real intention destination.
Therefore, in this paper, we focus on predicting users' intention destinations in online travel platforms.
In detail, we act as online travel platforms (such as Fliggy and Airbnb) to recommend travel plans for users, and the plan consists of various vacation items including hotel package, scenic packages and so on.
%
%In online travel platforms (e.g. Fliggy, Airbnb), vacation items (e.g. hotel package, scenic packages) are recommended for traveling users.
%And users may not be interested in a vacation item only if the item's location matches
%users' real intention destination.
%Thus, in this paper, we focus on studying users' intention destination prediction, 
%which is essential for recommender systems in online travel platforms.
%
%
Predicting the actual intention destination in travel planning is challenging.
Firstly, users' intention destination is highly related to their travel status (e.g., planning for a trip or finishing a trip).
%; moreover, users' latest orders in online travel platforms may not indicate his real intention.
Secondly, users' actions (e.g. clicking, searching) over different product types (e.g. train tickets, visa application) have different indications in destination prediction.
Thirdly, users may mostly visit the travel platforms just before public holidays, 
and thus user behaviors in online travel platforms are more sparse, low-frequency and long-period.
Therefore,  
we propose a Deep Multi-Sequences fused neural Networks (DMSN) to predict intention destinations from fused multi-behavior sequences. 
%Moreover,
%to capture users' long-term and impromptu preferences for intention destinations, as well as 
%learn the interactions among users' actions,   
%to emphasize the user preference for different destinations during different period, 
%we intersect the DMSN model with attention-based neural networks.
%
Real datasets are used to evaluate the performance of our proposed DMSN models.
%Besides, to verify the effectiveness of DMSN models in real applications, we merge them into existing click-through rate (CTR) prediction models.
Experimental results indicate that the proposed DMSN models can achieve high intention destination prediction accuracy.
% as well as assist to obtain better CTR prediction performance.
\end{abstract}


% \fi
\IEEEpeerreviewmaketitle

\section{Introduction} \label{sec:introduction}
Travel planning has attracted more and more attentions recent years and 
artificial neural networks are widely used to provide satisfactory plannings.
The quality of the recommended travel plan is affected by users'
transportation type, intention destination estimation,  budget limit and crowdness prediction. 
Among these factors, user' intention destination estimation is the most important and challenging factor since the user may not be interested in 
the recommended travel plan unless the plan matches the user's real intention destination. 

%ersonalized recommendation method, based on the intuition that users' interests can be inferred from their historical behaviours or other users with similar preference, has been widely used in many real-world applications~\cite{DBLP:conf/kdd/ZhuLZLHLG18}. Good personalized recommendations can enrich users' experiences, and thus e-commerce leaders like Taobao, Amazon and Netflix have made recommender systems a salient part of their platforms \cite{DBLP:journals/computer/KorenBV09}. 

\begin{figure}[!htbp]
	\centering
	\includegraphics[width=0.98\columnwidth]{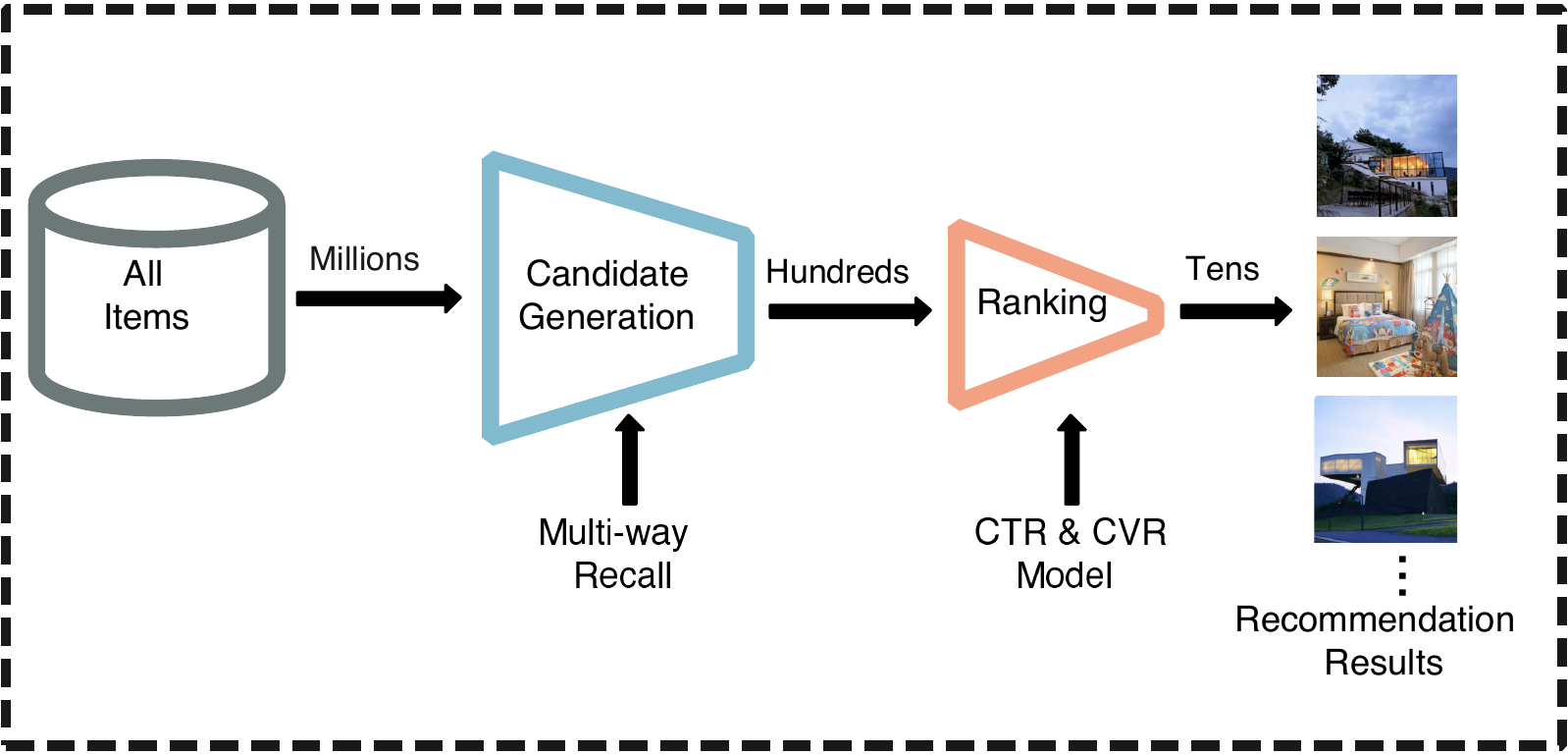}
	\caption{A general framework of online recommender system}\label{fig:RS-framework}
\end{figure}

To enrich users' experiences, intention prediction has been studied in many real e-commerce recommender systems like like Taobao, Amazon\cite{DBLP:conf/kdd/ZhuLZLHLG18, DBLP:journals/computer/KorenBV09}.
Online user intention prediction includes predictions of users' online clicking, searching and purchasing behaviors, etc.
%As illustrated in Figure~\ref{fig:RS-framework}, the part of dotted line shows a brief online recommended link of e-commerce recommender system, 
Figure~\ref{fig:RS-framework} illustrates a simple online recommending link in e-commerce recommender systems,
and the link consists of two stages: matching and ranking. 
During the matching stage, the recommender systems will generate hundreds of candidate items from item pool by multi-way recall such as the item-item recall (i2i), destination-item recall (d2i) and so on.
During the ranking stage, most recommender systems conduct click-through rate prediction~\cite{DBLP:conf/ijcai/GuoTYLH17,DBLP:conf/recsys/Cheng0HSCAACCIA16,DBLP:conf/kdd/ZhouZSFZMYJLG18,DBLP:conf/cikm/SongS0DX0T19, DBLP:conf/kdd/OuyangZLZXLD19}, and some also do a post-click conversion rate prediction~\cite{DBLP:conf/sigir/MaZHWHZG18,DBLP:conf/aaai/WenZLYH19,DBLP:conf/sigir/Shan0S18}. 
%CTR models~\cite{DBLP:conf/ijcai/GuoTYLH17,DBLP:conf/recsys/Cheng0HSCAACCIA16,DBLP:conf/kdd/ZhouZSFZMYJLG18,DBLP:conf/cikm/SongS0DX0T19, DBLP:conf/kdd/OuyangZLZXLD19} 
%mainly focus on estimating the probability a user will click on a recommended item, while CVR models \cite{DBLP:conf/sigir/MaZHWHZG18,DBLP:conf/aaai/WenZLYH19,DBLP:conf/sigir/Shan0S18} estimates the probability a user will purchase on a recommended item after clicking. 
%In general, the score of $ pCTR\ast pCVR $ is used to rank the items in candidate set and top ten-level items are selected as the final recommendation results to users in online recommendation platform. 
%In this paper, we focus on CTR predictions.

\begin{figure}[htbp]
	\centering
	\includegraphics[width=0.48\textwidth]{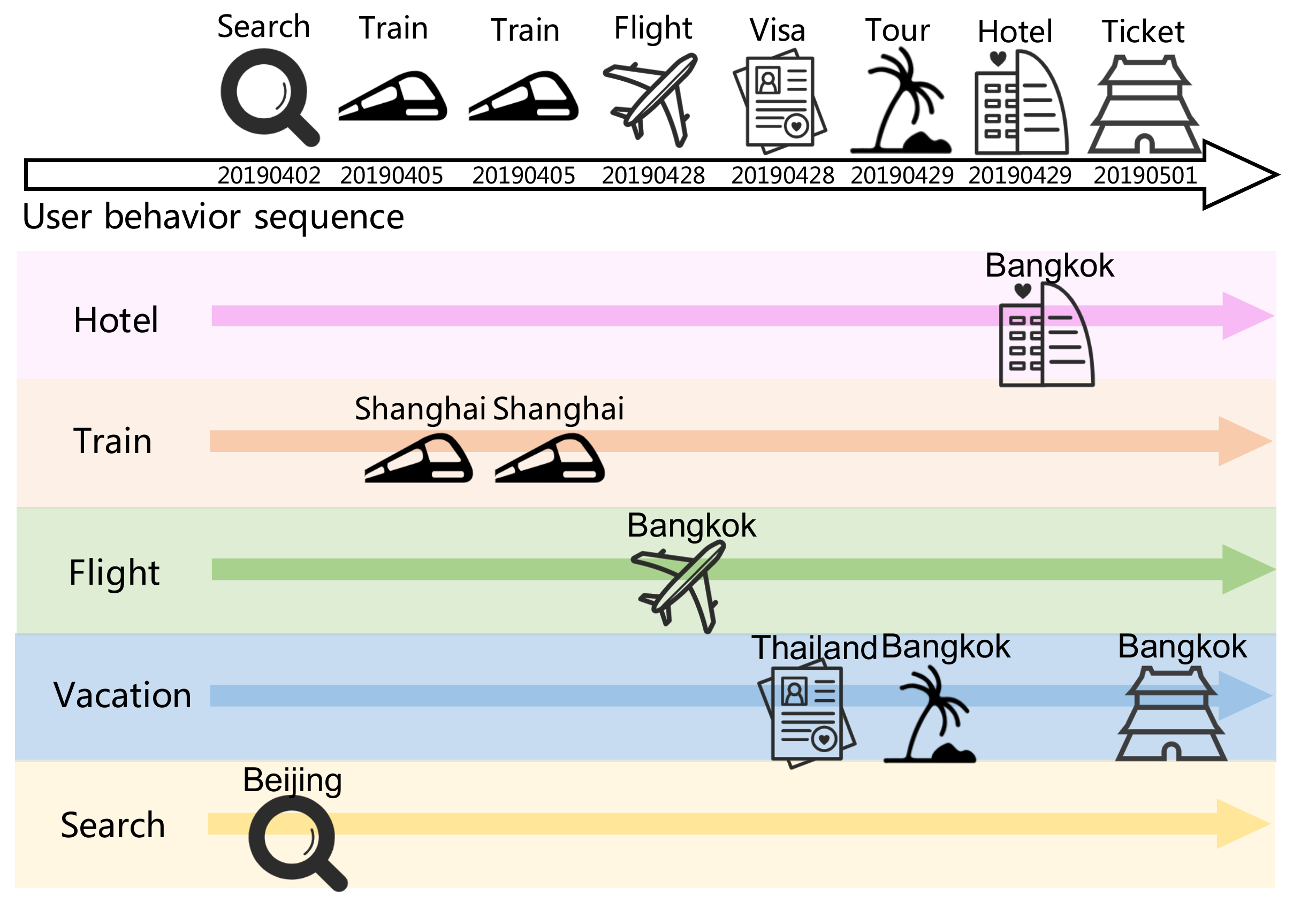}
	\caption{Example case in travel platforms}\label{fig:intro_example}
\end{figure}

Online travel platform has a key characteristic that distinguishes it from other e-commerce platforms (e.g. Google, Taobao, YouTube),
that is, users' behaviors and intentions are significantly related to {\em locations}.
Users' behaviors are composed by action types and product types, where
action types include {\em click, purchase, collect} and
product types include {\em hotel, train ticket, flight ticket, vacation item, solitary search}.
For instance, searching Beijing, booking a hotel in Bangkok, and clicking the visa application in Thailand are users' possible behaviors.
Among different product types, {\em vacation item} is the recommendation target in online travel systems,
and vendors in travel platforms provide various categories of vacation items, such as visa applications, scene packages.
All products in online travel platforms are associated with locations.
When online travel platforms recommend a candidate vacation item to a user, the locations associated to the user's historical behaviors should be taken into consideration.
As shown in Figure~\ref{fig:intro_example}, the user {\em searched} Beijing on April 2nd 2019, {\em booked} train tickets for April 5th to Shanghai, after that, he {\em compared} flight tickets for April 28th to Bangkok, {\em browsed} hotels and scenic tickets around May 1st in Bangkok. 
In this example, recommending a hotel package in Bangkok will be much better than a package in Shanghai.
The reason is that, according to his historical behaviors, the user only bought tickets for April 5th to Shanghai, then it is possible that he lives in ShangHai and just went back home after a business trip on April 5th.
In contrast, the user browsed various products frequently in Bangkok for the period from April 28th to May 1st,
and these behaviors indicate that the user may plan a trip to Bangkok during the public holiday around May 1st. 
Thus, recommending vacation items in Bangkok may match the user better. 
When recommending travel-related products to a user, we need to prioritize the accuracy of the recommended destination. Inspired by the above observation, we pay more attentions to those geo-features during vacation item recommendations, 
which has not been considered in most existing online intention prediction models. 
{\em In this paper, we focus on predicting users' intention destinations for online travel recommender systems.}

Predicting the actual destination is not easy as users' intention destinations are highly related to their travel status, and user's latest orders may not express his real intention. 
For instance, the latest purchasing order made by the user in Figure~\ref{fig:intro_example} is booking the train tickets to Shanghai, but Shanghai may not be his real intention destination as analyzed above.
Moreover, the intersections between users' behaviors may give more precise indications of intention destination.
For instance, the user searched/browsed vacation items of different product types in Bangkok in Figure~\ref{fig:intro_example} may indicate that he is really interested in Bangkok.

Despite {\em understanding users' travel status} and 
{\em figuring out the intersections between users' behaviors}, 
another challenge in intention destination prediction is that,
 users' historical behaviors are quite sparse, low-frequency and long period in online travel platforms.
Compared with Google, Youtube and Taobao, users may only visit the online travel platforms during public holidays,
and leave the platforms for several months after a trip, which make the behavior data low-frequency and long period.
%Moreover, users may only buy train or flight tickets on the platforms, leaving no intention information for future recommendation. 
Besides, 
as categorical features are represented with one-hot encoding,
 huge number of different products in online travel platforms makes
historical behavior features sparse and high-dimensional.
Moreover, users' behaviors can infer his {\em long-term} and {\em short-term} preferences.
For instance, a user may have visited Japan twice a year for several years, 
inferring that his {\em long-term} preference is Japan; 
while, he may search visa application services in Thailand and Phra Phrom recent days,
inferring that his {\em short-term} preference is Thailand.
Thus, \textit{how to enrich and emphasize users' features according to their historical behaviors for intention destination prediction} is challenging and urgent.

\begin{figure}[htbp]
	\centering
	\includegraphics[width=0.48\textwidth]{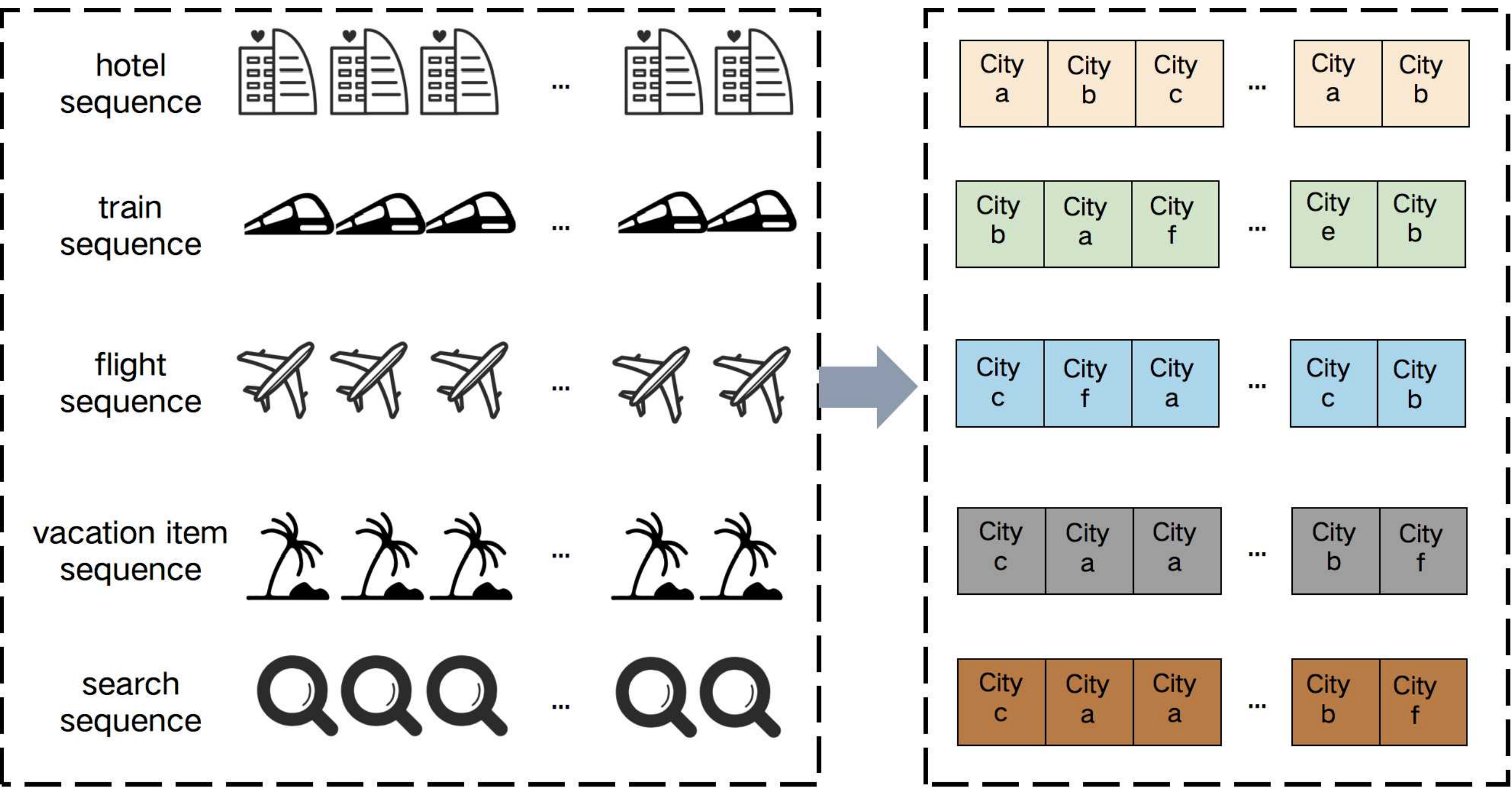}
	\caption{A mapping from five historical behavior sequences to cityIDs sequences}\label{fig:transform}
\end{figure}

To predict users' intention destinations over limited historical data, we first propose a framework called Deep Multi-Sequences fused neural Networks (denoted as DMSN) to make fully effective use of user's behaviors over different product types. Considering the all user behaviors on the online travel platform are related to locations, we first map a user's behaviors over different product types into multiple cityID sequences, as depicted in Figure~\ref{fig:transform}. In order to make fully effective use of the correlation between the user's multiple behavior sequences, we fuse and align the user's multiple behavior sequences into a global cityID sequence in the order of behavior time to predict users' real intention destinations. When a user has no behavior on some product types, we will not consider the corresponding behavior types in the fusion sequence. But we will use a default behavior sequence of length one to represent the behavior type in which the user has no behavior if we model multiple behavior sequences separately. Moreover,using cityIDs can reduce the sparsity of categorical behavior features as there are tens of thousands of cities in the whole world and popular tourist cities are only hundreds while there are millions of different products in the platforms.
% To predict users' intention destinations over limited historical data,
% %we propose a neural network to predict users' intention destinations from their historical behaviors. 
% %
% we first propose a framework called Deep Multi-Squences fused neural Networks (denoted as DMSN) to enrich users' features.
% That is, we fuse and map a user's behaviors over different product types into a global cityID sequence. 
% As depicted in Figure~\ref{fig:transform}, as each product is associated with locations,
% we map a user's behavior sequence into a cityID sequence in the order of behavior time for each product type,
% and then we fuse those five cityID sequences into a global sequence to predict users' real intention destinations. 
% Using cityIDs can reduce the sparsity of categorical behavior features as there are only hundreds of cities in the whole world while
% there are millions of different products in the platforms. 
%
%
%
%Secondly, as discussed before, 
As users' intention destinations are highly related to their travel status, which may change with time, we use behavior {\em time} as an indicator to calculate attention weights of different historical behaviors. We intersect Recurrent Neural Network and Multi-grained Convolutional Neural Network into DMSN model, and come up with two models: ATtention based Recurrent Neural Network model (DMSN-ATRNN) and ATtention based Multi-grained Convolutional neural network model (DMSN-ATMC). DMSN-ATRNN can effectively capture the change of users' long-term preference for different destinations, but it is limited on the long training process and poor performance in mining the pattern of short-term behavior. By extracting features with multi-grained convolution neural network, DMSN-ATMC can capture user's short-term and long-term preferences simultaneously and thus can achieve high intention destination prediction accuracy.
% With enriched user features, we predict users' intention destinations.
% As users' intention destinations are highly related to their travel status, which may change with time,
% %Thus, in order to predict users' intention destinations accurately, 
% we use behavior {\em time} as an indicator to calculate attention weights of different historical behaviors.
% %
% We intersect Recurrent Neural Network and Multi-grained Convolutional Neural Network into DMSN model, 
% and come up with two models: ATtention based Recurrent Neural Network model (DMSN-ATRNN) and ATtention based Multi-grained Convolutional neural network model (DMSN-ATMC).
% DMSN-ATRNN can effectively capture the change of users' long-term preference for different destinations, 
% but it is limited on the long training process and poor performance in mining the pattern of short-term behavior. 
% By extracting features with multi-grained convolution neural network, DMSN-ATMC can capture user's short-term and long-term preferences simultaneously and thus can achieve high intention destination prediction accuracy.

Moreover, our proposed DMSN models are generalized models, which can be used in different business applications through merging into different existing intention prediction models (e.g. CTR models DIN~\cite{DBLP:conf/kdd/ZhouZSFZMYJLG18}, AutoInt~\cite{DBLP:conf/cikm/SongS0DX0T19}).
When merging the DMSN model into existing CTR prediction models~\cite{DBLP:conf/kdd/ZhouZSFZMYJLG18,DBLP:conf/cikm/SongS0DX0T19}, we can achieve: i) users' preference over different cities can be learned from the global behavior features and other auxiliary features and ii) the accuracy of the CTR predictions in online travel platforms can be improved with the use of predicted intention destinations.

%\footnote{\url{https://proceedings.ijcai.org/info}}
In summary, we make following contributions:
\begin{itemize}
%\item A new modeling process of CTR prediction in online travel platforms, i.e., $context\to destination\to vacation~item$, is proposed. 
\item A generalized intention destination prediction framework (i.e., DMSN) for online travel recommender systems is proposed.
\item Attention-based neural networks are integrated into DMSN models to effectively capture the change of users' long-term and short-term preferences.
%\item Comprehensive experiments on real log datasets and online A/B testing are conducted to evaluate the effectiveness of proposed models.
\item Comprehensive experiments on real log datasets are conducted to evaluate the effectiveness of proposed models.
\item Code and a large scale of high quality log dataset of travel scenarios will be released. The dataset will contribute to the research of personalized recommendation in travel scenarios.
\end{itemize}

The rest of this paper is organized as follows: 
%Sections~\ref{sec:problem} describe the studied problem.
Section~\ref{sec:method} describes the proposed models in detail and Section~\ref{sec:experiment} illustrates the experimental results.
Section~\ref{sec:related} discusses related literatures and Section~\ref{sec:conclude} concludes this paper.
%Section 2 describe the proposed model in detail. Section 3 and Section 4 focus on the experimental results about the proposed model, including offline evaluation and online A/B test performance. At last, we conclude the paper in Section 5.

\section{Problem Statement and System Architecture} \label{sec:problem}
In online travel platforms, online intention prediction aims to predict the probability of a user clicking/searching/purchasing a vacation item according to his historical behavior context. 
%The problem of click-through rate (CTR) prediction aims to predict the probability of a user clicking on a vacation item according to his historical behavior context.
As discussed in Section~\ref{sec:introduction}, to provide more precise intention predictions over vacation items in online travel platforms, it is important to predict users' intention destination accurately. 
%However, most modeling process of current CTR models have not paid sufficient attention to the locations. 
Therefore, in this paper, we focus on predicting users' intention destinations using users' historical data.
%To compare, we propose a new modeling process of CTR prediction as $context\to destination\to vacation~item$ to make the recommended items more consistent with users' travel intentions.
%
\begin{problem}
Given a user's historical behavior sequences and a candidate intention destination $dest$, 
our goal is to predict the probability that the user will be interested in vacation items in $dest$.
\end{problem}

We propose Deep Multi-Squences fused neural Network (denoted as DMSN) models (Figures~\ref{fig:TARNN} and~\ref{fig:ATMC}) to predict users' real intention destinations in online travel platforms.
Our model follows DIN~\cite{DBLP:conf/kdd/ZhouZSFZMYJLG18} and shares a similar Embedding$\&$MLP paradigm as most of the popular model structures~\cite{DBLP:conf/kdd/ZhouZSFZMYJLG18,DBLP:conf/recsys/Cheng0HSCAACCIA16,paper2cite20}.
%
%The main goal of our approach is to predicate the users' travel intention, that is, which city the user will be more interested in?
We utilize users' historical behaviors to predicate user's preference score over a candidate item city.
As users' behaviors in travel platforms are quite sparse,
in the {\em Input Layer},
we map and fuse all five categories of users' historical behaviors into a global cityID sequence.
As data features in CTR predictions are mostly sparse and high-dimensional,
{\em Embedding Layer} is necessary to represent those high-dimensional sparse features into low-dimensional spaces.
Behaviors related to displayed ads greatly contribute to the click action.
A main characteristic of travel ads recommendation is that, 
users' interests on different cityIDs change with their travel periods. 
Thus, {\em Attention-based Neural Network Layer} (i.e., Attention layer and RNN layer in Figures~\ref{fig:TARNN}; Attention layer and CNN layer in ~\ref{fig:ATMC}) is applied to emphasize the user preference for different destinations during different time.
Moreover, to capture users' long-term and short-term preferences for different destinations, we use RNN and multi-grained CNN to express users' diverse interests.
Our model also follows DIN~\cite{DBLP:conf/kdd/ZhouZSFZMYJLG18} to adaptively calculate the representation vectors of user interests by taking into consideration the relevances between historical behaviors and the candidate cityID.
That is, in the {\em MLP layer}, the candidate item cityID is included in a fully connected layer to learn the combination of features automatically. 
The output layer {\em Rank Loss Layer} is used to finally calculate the user's preference score for the given candidate item cityID.

Compared with the existing intention prediction models~\cite{DBLP:conf/kdd/ZhouZSFZMYJLG18, DBLP:conf/cikm/SongS0DX0T19}, 
our model has three main differences: 
i) as in travel scenario, users' behaviors on vacation items are quite sparse, thus in the input layer, we map and fuse all five historical behavior sequences to achieve dense representations;
ii) users' interest over a city is highly affected by their travel status, i.e., the time period during their traveling, thus, we add time-factors in our attention functions;
iii) users' intention destinations can be different in long-term and short-term, thus we utilize RNN and multi-grained CNN models to better capture users' preferences. 
Details of our models will be discussed later in Sections~\ref{sec:method}.

\section{System Architecture} \label{sec:system}

\up
\begin{figure}[htbp]
	\centering
	\includegraphics[width=0.43\textwidth]{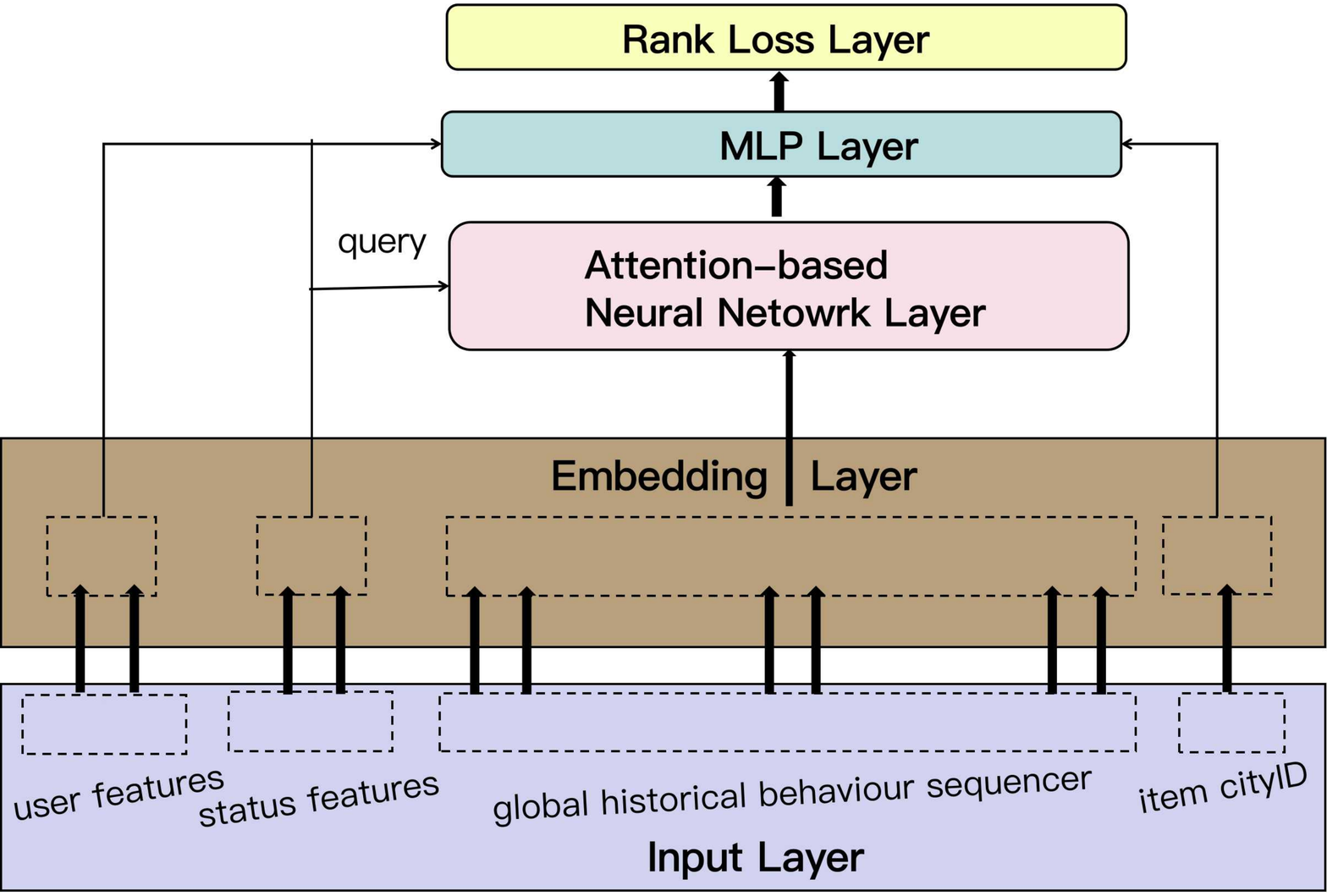}
	\caption{System architecture}\label{fig:sys}
\end{figure}
\down

This paper objects to predict users' real intention destinations in online travel platforms,
and 
Figure~\ref{fig:sys} illustrates the overall architecture of our proposed models.
Our model follows DIN~\cite{DBLP:conf/kdd/ZhouZSFZMYJLG18} and shares a similar Embedding$\&$MLP paradigm as most of the popular model structures~\cite{DBLP:conf/kdd/ZhouZSFZMYJLG18,DBLP:conf/recsys/Cheng0HSCAACCIA16,paper2cite20}.
%
%The main goal of our approach is to predicate the users' travel intention, that is, which city the user will be more interested in?
We utilize users' historical behaviors to predicate user's preference score over a candidate item city.
As users' behaviors in travel platforms are quite sparse,
in the {\em Input Layer},
we map and fuse all five categories of users' historical behaviors into a global cityID sequence.
As data features in CTR predictions are mostly sparse and high-dimensional,
{\em Embedding Layer} is necessary to represent those high-dimensional sparse features into low-dimensional spaces.
Behaviors related to displayed ads greatly contribute to the click action.
A main characteristic of travel ads recommendation is that, 
users' interests on different cityIDs change with their travel periods. 
Thus, {\em Attention-based Neural Network Layer} is applied to emphasize the user preference for different destinations during different time.
Moreover, to capture users' long-term and short-term preferences for different destinations, we use RNN and multi-grained CNN to express users' diverse interests.
Our model also follows DIN~\cite{DBLP:conf/kdd/ZhouZSFZMYJLG18} to adaptively calculate the representation vectors of user interests by taking into consideration the relevances between historical behaviors and the candidate cityID.
That is, in the {\em MLP layer}, the candidate item cityID is included in a fully connected layer to learn the combination of features automatically. 
The output layer {\em Rank Loss Layer} is used to finally calculate the user's preference score for the given candidate item cityID.

Compared with existing CTR prediction models, 
our model have three main differences: 
i) as in travel scenario, users' behaviors on vacation items are quite sparse, thus in the input layer, we map and fuse all five historical behavior sequences to achieve dense representations;
ii) users' interest over a city is highly affected by their travel status, i.e., the time period during their traveling, thus, we add time-factors in our attention functions;
iii) users' intention destinations can be different in long-term and short-term, thus we utilize RNN and multi-grained CNN models to better capture users' preferences. 
Details of our models will be discussed later in Sections~\ref{sec:method}.

\section{DMSN models} \label{sec:method}
In this section, we describe our proposed Deep Multi-Squences fused neural Networks (denoted as DMSN) models.
Given a user's historical behavior sequences and a candidate intention destination, 
the goal of DMSN models is to predict the probability that the user 
will be interested in vacation items in the candidate destination.

\subsection{DMSN Overview}
As depicted in Figures~\ref{fig:TARNN} and~\ref{fig:ATMC},
our model follows DIN~\cite{DBLP:conf/kdd/ZhouZSFZMYJLG18} and shares a similar Embedding$\&$MLP paradigm as most of the popular model structures~\cite{DBLP:conf/kdd/ZhouZSFZMYJLG18,DBLP:conf/recsys/Cheng0HSCAACCIA16,paper2cite20}.
We utilize users' historical behaviors to predicate user's preference score over a candidate item city.
As users' behaviors in travel platforms are quite sparse,
we map and fuse all five categories of users' historical behaviors into a global cityID sequence,
and the global cityID sequence is used as the {\em Input}.
As data features in intention destination predictions are mostly sparse and high-dimensional,
{\em Embedding Layer} is necessary to represent those high-dimensional sparse features into low-dimensional spaces.
A main characteristic of online travel platforms is that, 
users' interests on different cityIDs change with their travel periods. 
Thus, {\em Attention-based Neural Networks} is applied to emphasize the user preference for different destinations during different time.
In detail, to capture users' long and short-term preferences for different destinations, we use RNN and multi-grained CNN to express users' diverse interests (i.e., Attention layer and RNN layer in Figure~\ref{fig:TARNN}; Attention layer and CNN layer in Figure~\ref{fig:ATMC}).
Our models also adaptively calculate the representation vectors of user interests by taking into consideration the relevances between historical behaviors and the candidate cityID.
That is, in the {\em MLP layer}, the candidate item cityID is included in a fully connected layer to learn the combination of features automatically. 
The output of DMSN models is the user's preference score for the given candidate item cityID.

Compared with existing intention prediction models~\cite{DBLP:conf/kdd/ZhouZSFZMYJLG18, DBLP:conf/cikm/SongS0DX0T19}, 
DMSN models have three main differences: 
i) as in travel scenario, users' behaviors on vacation items are quite sparse, thus in the input, we map and fuse all five historical behavior sequences to achieve dense representations;
ii) users' interest over a city is highly affected by their travel status, i.e., the time period during their traveling, thus, we add time-factors in our attention functions;
iii) users' intention destinations may be different in long-term and short-term, thus RNN and multi-grained CNN are used to better capture users' preferences. 

In the following of this section, we discuss the proposed DMSN-ATRNN and DMSN-ATMC models in detail.

\subsection{DMSN-ATRNN Model}
\begin{figure*}[htbp]
	\centering
	\includegraphics[width=0.85\textwidth]{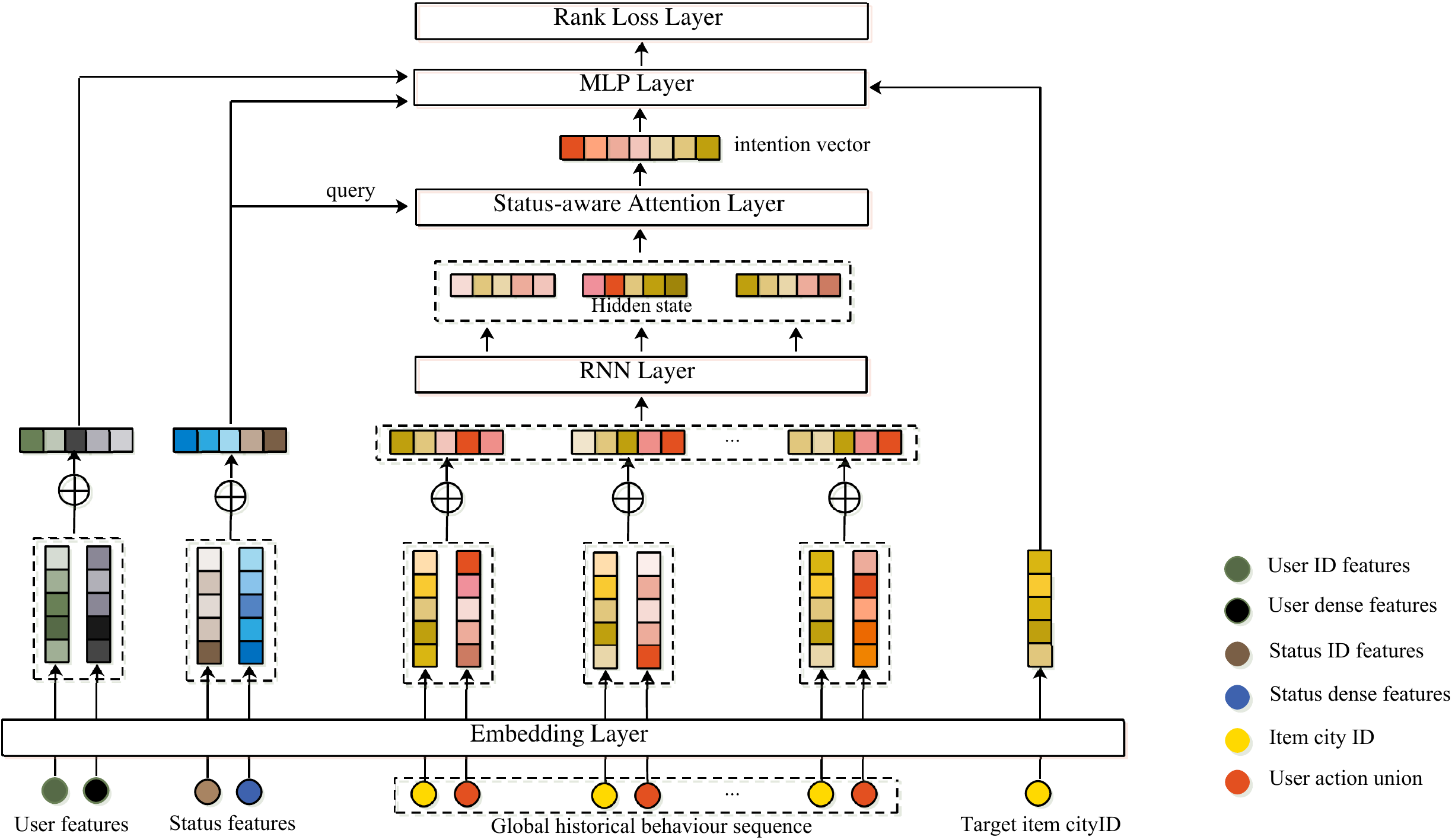}
	\caption{Architecture overview of DMSN-ATRNN}\label{fig:TARNN}
\end{figure*}

In DMSN-ATRNN model, we use recurrent neural network in the Attention-based Neural Network layer to capture the sequential correlation of global behavior sequence and weight historical behaviors.
%Because, different behaviour types of user are complementary in travel destination prediction, we fuse all five types of behaviours sequences into a global sequence according to global behaviour time. Moreover, attention mechanism is implemented in ATRNN to weight historical behaviour.
Figure~\ref{fig:TARNN} illustrates the architecture of the DMSN-ATRNN model.
In the rest of this section, we discuss each layer in DMSN-ATRNN, and we use ATRNN for short.

\subsubsection{Input Layer}\label{sec:inputlayer}
There are four components in the Input Layer.

\stitle{1) User features} User features contain the basic features of users, including `user ID', `user age', `gender', `purchase-level' and so on. 
We denote the set of \underline{u}ser features as $\left \{ x_{ui} \right \}_{i=1}^{n_{u}}$, where $n_{u}$ is the number of user features in travel platforms.

\stitle{2) Status features} User's status information is very important in the travel scenario, for instance, recommending visa application services in the beginning of the trip, scenic packages in the middle of trip and flight tickets in the end of the trip will achieve better click-through rates. 
We obtain status features from users' itinerary, which contain information like the destination of the trip, the departure time of trip, the completion status of the trip and so on. In addition to user's status information, time-related information is also taken into account, such as the current month, season, hour, etc.
We denote the status features obtained from \underline{o}rders as $\left \{ x_{oi} \right \}_{i=1}^{n_{o}}$, where $n_{o}$ is the dimension of status features.

\stitle{3) Sequential features} 
As discussed in Figure~\ref{fig:transform} in Section~\ref{sec:introduction}, in order to utilize all historical behavior information of a user, we map and fuse different behavior sequences into a global cityID sequence. 
%Five behaviour sequences are fused into the sequential feature according to the global behavior time. 
Each behavior sequence consists of a series of `action instance', and each `action instance' contains `city ID' and `action unit'. 
The `action unit' is a {\em Cartesian Product} of two sets: product type set $P$ and action type set $A$,
where $P$ = \{train ticket, flight ticket, vacation item, hotel, search\} and $A$ = \{click, purchase, collect\}. 
Moreover,  each `action unit' will be discretized to ID feature before transferred to the Embedding Layer. 
The fused global cityID sequence put all `action instances' in the order of corresponding behavior time.
We denote the set of \underline{s}equential features as $\left \{ x_{si} \right \}_{i=1}^{n_{s}}$, where $x_{si}=[x_{cityID};x_{action~unit}]_{si}$, and $n_{s}$ is the length of sequential features.
As the sequential features are represented as one-hot encoding and
the number of different cities is significantly smaller than that of different products,
using cityID instead of specific product can reduce the sparsity of sequential features. 
Moreover, fusing all behaviors into a single cityID sequence can help relax the problem of low-frequency and long period.

%There are two fusing timings in our recommendation model:
%one is to fuse all features in the very beginning as a global sequence while 
%the other is to process each behavior sequence independently and fuse them before full concatenation in MLP layer.
%We compare these two strategies in the experiment and choose the better one, i.e., fusing in the beginning, to use in our model.  

\stitle{4) Item features} 
The output of ATRNN model is the predicted preference score of a given candidate city, and
thus the candidate `city ID' is used as the item feature. We denote $x_{i}$ as the item feature.

\subsubsection{Embedding Layer}
In recommender systems, the input features are sparse and have huge dimension, different from computer vision. An embedding is a mapping of a discrete-categorical-variable to a dense vector of continuous numbers, which has been widely used in Natural Language Processing (NLP) \cite{DBLP:conf/nips/MikolovSCCD13} and Recommender System (RS) to alleviate the above phenomenon. In ATRNN, all input features are mapped into the dense vector after embedding, i.e., $\left \{ \mathbf{X}_{ui} \right \}_{i=1}^{n_{u}}$, $\left \{ \mathbf{X}_{oi} \right \}_{i=1}^{n_{o}}$, $\left \{ \mathbf{X}_{si} \right \}_{i=1}^{n_{s}}$ and $\mathbf{X}_{i}$, which contain richer useful information and yield better generlization. 

\subsubsection{Status-aware Attention Layer}
Attention mechanism~\cite{DBLP:conf/iclr/KimDHR17,DBLP:conf/nips/VaswaniSPUJGKP17} is firstly introduced in the encoder-decoder framework for machine translation systems, and it allows the model to emphasize the effect of relevant parts in the input sequence as needed. 
% 
%layer is applied over a RNN layer to emphasize the user preferences for different destinations at different times. 
In this section, we denote $t$ as the current time, $t_{i}$ as the timestamp $i$, $\mathbf{X}_{o}$ as the concatenation vector of all embedding vector of status features, $\mathbf{\bar{h}}_{i}$ as the hidden state of RNN layer at time $i$, $\mathbf{W}_{a}$ and $\mathbf{W}_{b}$ as the learnable parameters in the attention layer, and set $\tau _{i}=t_{i}-t$. 
In detail, the implementation of attention for sequence-to-one networks is shown in Eq.~\ref{eq:attention1}, Eq.~\ref{eq:attention2}, Eq.~\ref{eq:attention3} and Eq.~\ref{eq:attention4}. 
Attention weights based on status information are calculated using $\mathbf{X}_{o}$ as shown in Eq.~\ref{eq:attention1}:

\begin{equation} 
\begin{split}
\alpha _{ti}=\frac{exp(score(\mathbf{X}_{o},\left [ \mathbf{\bar{h}}_{i}, \mathbf{T_{i}}\right ]))}{\sum_{i^{'}=1}^{T}exp(score(\mathbf{X}_{o},\left [ \mathbf{\bar{h}}_{i}, \mathbf{T_{i}}\right ]))}
\end{split} \label{eq:attention1}
\end{equation}
where:
\begin{equation}
\begin{split}
score(\mathbf{X}_{o},\left [ \mathbf{\bar{h}}_{i}, \mathbf{T_{i}}\right ])=\mathbf{X}_{o}^{T}\mathbf{W}_{a}\left [ \mathbf{\bar{h}}_{i},\mathbf{T_{i}}\right ]
\end{split} \label{eq:attention2}
\end{equation}
% time attention weights are calculated in Eq.(6):
% \begin{equation}
% \begin{split}
% att_{ti}=\frac{exp(\alpha _{ti}\times \tau _{i})}{\sum_{i^{'}=1}^{T}exp(\alpha _{ti^{'}}\times \tau _{\tau^{'}})}
% \end{split}
% \end{equation}
and
\begin{equation}
\begin{split}
 \mathbf{T_{i}} = tanh(\mathbf{W_{b}} * log(1+|\tau _{i}|))
\end{split} \label{eq:attention3}
\end{equation}
The output of Status-aware Attention Layer 
is an intention vector $\mathbf{a}_{t}$ calculated using Eq.~\ref{eq:attention4}.
$\mathbf{a}_{t}$ is computed as a weighted sum of the $\left \{ \mathbf{\bar{h}}_{i} \right \}_{i=1}^{n_{s}}$,
where the weight assigned to each $\mathbf{\bar{h}}_{i}$ is computed by a function of the status features $\mathbf{X}_{o}$.
\begin{equation}
\begin{split}
\mathbf{a}_{t}=\sum_{i=1}^{t}\alpha _{ti}\times \mathbf{\bar{h}}_{i}
\end{split} \label{eq:attention4}
\end{equation}

\begin{figure*}[htbp]
	\centering
	\includegraphics[width=0.75\textwidth]{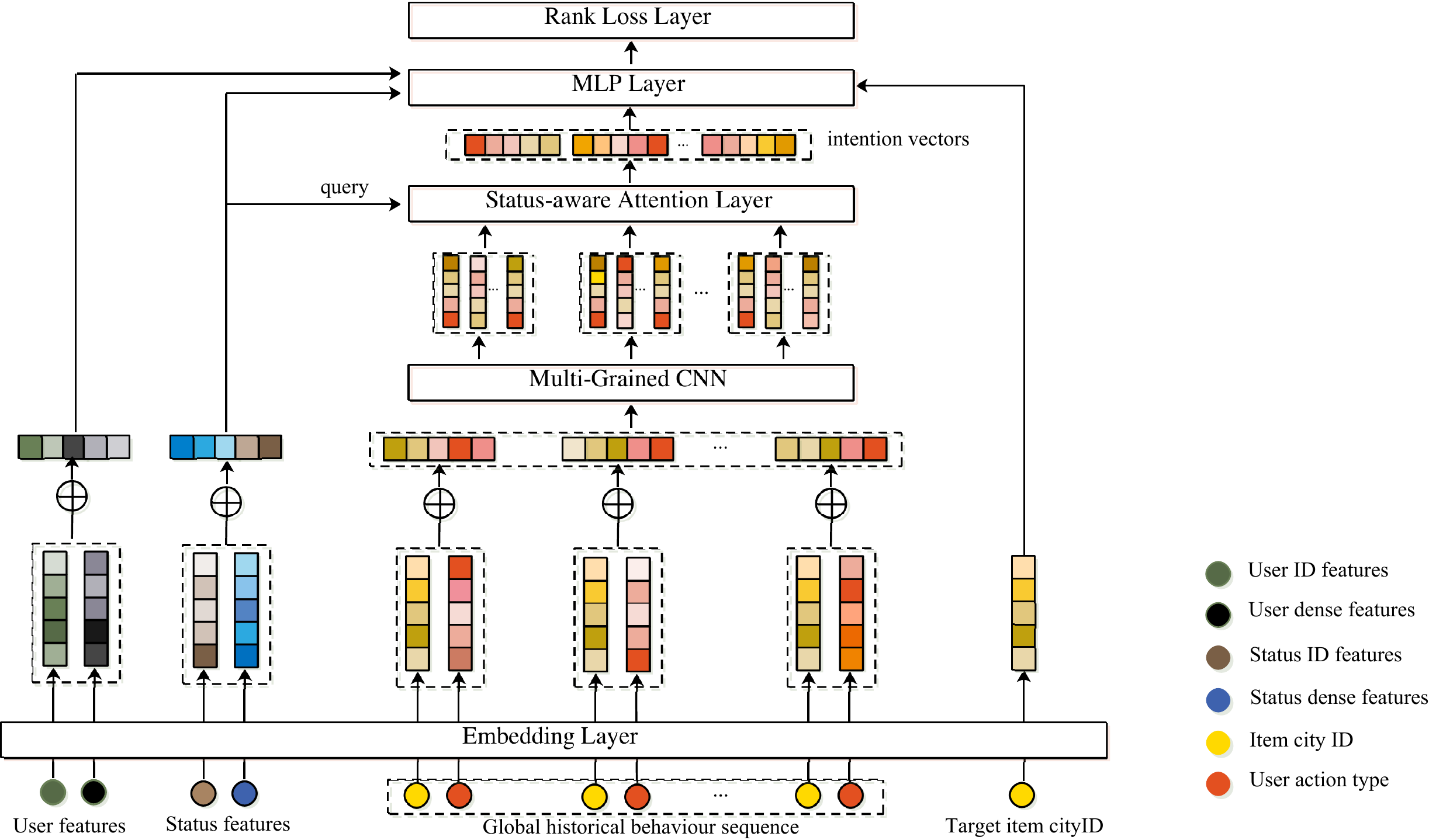}
	\caption{Architecture overview of ATMC}\label{fig:ATMC}
\end{figure*}

\subsubsection{MLP Layer}
The MultiLayer Perceptron (MLP) layer is a feed-forward neural network, which can generalize better to unseen feature combinations through low-dimensional dense embeddings learned for the sparse features \cite{DBLP:conf/recsys/Cheng0HSCAACCIA16}. We denote the input of MLP layer as $\mathbf{V}$, the output of MLP layer as $\mathbf{X}_{mlp}$. $\mathbf{V}=[\mathbf{X}_{u},\mathbf{X}_{o},\mathbf{a}_{t},\mathbf{X}_{i}]$, $\mathbf{X}_{u}$ is the concatenation vectors of all embedding vectors in $\left \{ \mathbf{X}_{ui} \right \}_{i=1}^{n_{u}}$, $\mathbf{X}_{o}$ is the concatenation vector of all embedding vectors in $\left \{ \mathbf{X}_{oi} \right \}_{i=1}^{n_{o}}$ and $\mathbf{a}_{t}$ is the output of attention layer.

\subsubsection{Rank Loss Layer}
The output of ATRNN is the preferences score of $user_{j}$ for the item's cityID, in this paper, we model this process as a point-wise ranking problem. We denote $\widehat{y}$ as the output of rank loss layer, meanwhile, $\widehat{y}$ is also the preferences score. $\mathbf{W}_{r}$ is the learnable parameters in rank loss layer. $\mathbf{X}_{mlp}$ is the output of MLP layer.
\begin{equation}
\begin{split}
\widehat{y}=\frac{1}{1+exp(-\mathbf{W}_{r}^{T}\mathbf{X}_{mlp})}
\end{split}
\end{equation}

$y$ is binary labels with $y=1$ or $y=0$ indicating whether click or not. The logistic loss of ATRNN is shown in Eq.(9):
\begin{equation}
\begin{split}
L(y,\widehat{y})=-ylog(\widehat{y})-(1-y)log(1-\widehat{y})
\end{split}
\end{equation}

Based on the characteristic of RNN, 
DMSN-ATRNN can effectively capture the change of users' long-term preference for different destination,
and this had been verified in a visualized real example in Figures~\ref{fig:visualcase}a,b.
However, ATRNN still has the following limitations: 
1)The users' short-term preference changes can not be well learned;
2) RNN models become difficult to train when the length of sequence is too long.

\subsection{DMSN-ATMC Model}\label{subsec:dmsnmodel}
As discussed above, DMSN-ATRNN can capture long-term preference well, but for short-term preferences, it cannot handle well.
For instance, if the user has no behaviors related to New York before, and just start to click/purchase tickets/vacation packages in New York during last few minutes. Then, ATRNN may not be able to capture the importance of features related to New York and the short-term preferences will be missed.
To solve this problem, we propose the DMSN-ATMC model, which intersects \underline{at}tenion based \underline{m}ulti-grained \underline{c}onvolutional neural network into the DMSN model.
By extracting features with multi-grained convolution kernels, ATMC can capture user's short-term and long-term preferences simultaneously. 
Compared with DMSN-ATRNN, the ATMC model uses multi-grained CNN layer instead of RNN layer as shown in Figure~\ref{fig:ATMC}. Different sizes of convolutional kernels are used to capture long-term and short-term preferences.

%Different from RNNs model, the training process of CNNs model is faster and easier to converge. 
%Figure~\ref{fig:ATMC} depicts the architecture of ATMC and we can see that ATMC is the same with ATRNN except the multi-grained CNN layer.
%

\begin{figure}[!htbp]
	\centering
	\includegraphics[width=0.5\textwidth]{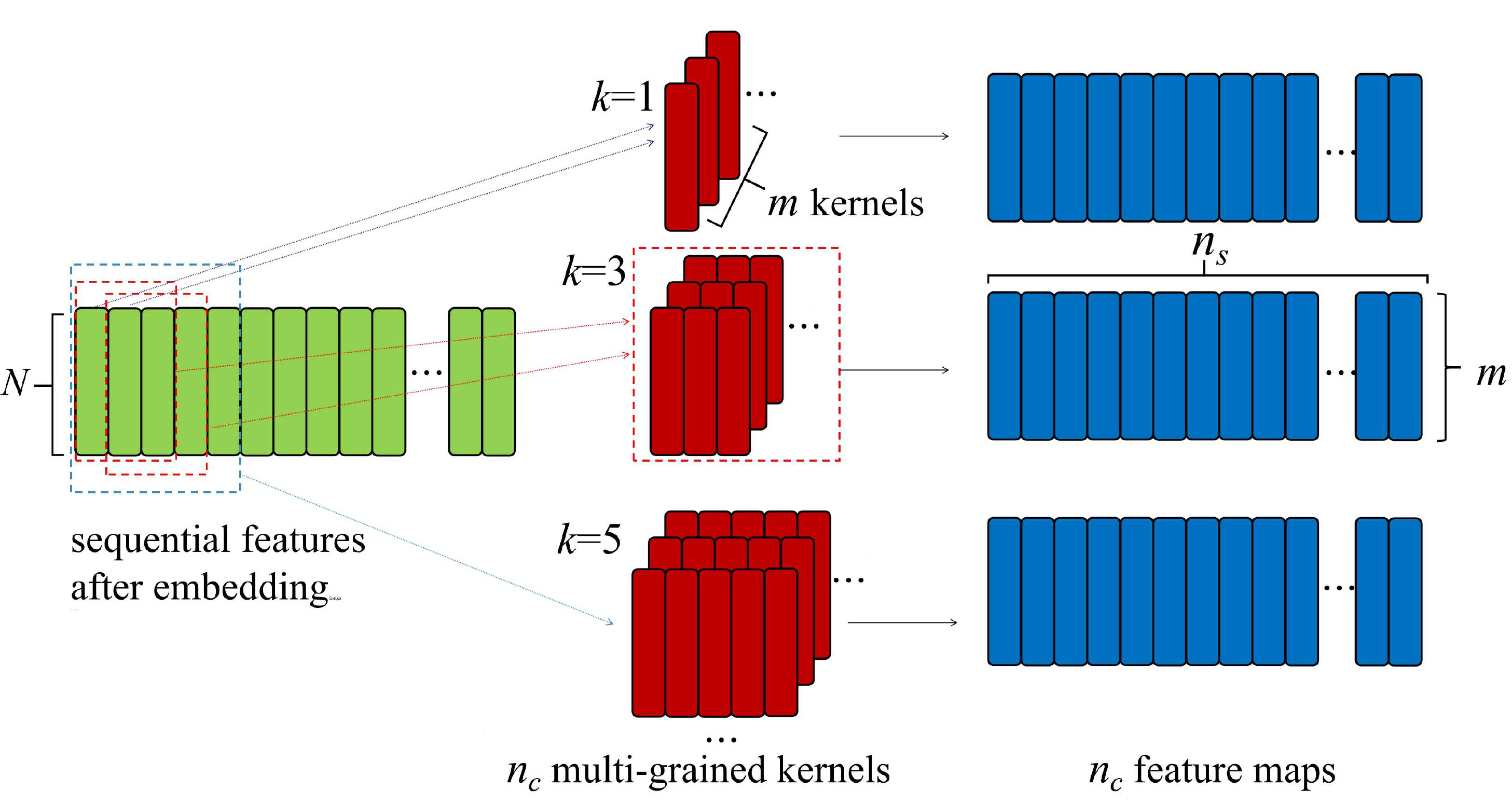}
	\caption{Architecture overview of multi-grained CNN. Multi-grained convolution kernels are used to extract feature for sequential features and generate multiple feature maps.}\label{fig:MGCNN}
\end{figure}

%\subsubsection{Multi-Grained CNN Layer}~\label{method:mgcnn}
The multi-grained CNN layer is 
inspired by the multi-grained scanning in multi-Grained Cascade Forest (gcForest) \cite{DBLP:conf/ijcai/ZhouF17}.
Multi-grained CNN layer can capture the feature representation in different time span.
%The architecture of multi-grained CNN illustrated in Fig.\ref{fig:ATMC}.
%
Fig.\ref{fig:MGCNN} shows three sizes of convolution kernels. We denote the dimension of embedding vector as $N$, the number of kernels of each granularity as $m$, the number of differently grained convolution kernel as $n_{c}$, the shape of sequential features as $N\times n_{s}$ and the shape of convolution kernel as $N\times k$. By using multiple size of convolution kernels (padding = same), different feature maps are generated. The shape of feature maps is $n_{c}\times m\times n_{s}$. It is observed that differently convolution kernels can cover different range of behaviour sequences. Note that the smaller the $k$, the more attention on user's short-term behaviour, and the larger the $k$, the more attention on user's long-term behaviour. Therefore, multi-grained convolution kernels can capture user's short-term and long-term preferences simultaneously.
We can get $n_{c}$ feature maps from the output of multi-grained CNN layer and each feature map is passed into the attention layer independently. The output of attention layer are $n_{c}$ attention vectors, which are denoted as $\left \{ \mathbf{a} _{ti}\right \}_{i=1}^{n_{c}}$. The concatenation of $\left \{ \mathbf{a} _{ti}\right \}_{i=1}^{n_{c}}$ as the final attention vector input to the MLP layer.

\section{Experimental Evaluation} \label{sec:experiment}

In this section, experiments are conducted over Fliggy's real data to evaluate the performance of proposed models. 
%We first describe the experiment settings, then we 
%show a visualized case study on attention weights obtained in our models to indicate the effectiveness of proposed time-aided attention functions.
%After that, we use log datasets to test the robustness and accuracy of our prediction models, 
%here we compare the accuracy in terms of both intention destination prediction and vacation item prediction.
%In the end, we integrate our models into a real online travel platform to test their online performances. 

%In this section, we conduct experiments on Fliggy's offline log dataset to evaluate the performance of different DMSN models and some classical CTR prediction models.

\subsection{Experiment Setup}

\subsubsection{Datasets}
In the experiments, we use real data from Fliggy
\footnote{https://www.fliggy.com/}, a well-known online travel recommendation platform.
We collect online log data from Fliggy's recommender system in Dec. 2019.
Fliggy has four business lines: train tickets booking, flight tickets booking, hotel booking and vacation item purchasing.
Relevant vacation items will be recommended to a user after he makes an order.
For instance, after a user buys a flight ticket to Bangkok, visa application services in Thailand may be recommended to him.  
%During our survey, no public datasets with multi-behaviour sequences for CTR prediction in travel scenarios. To evaluate the proposed approach, we collect the offline log data from Fliggy's recommender system in the past one month.

% Different behaviours data contain multiple action types of users, such as, click, purchase and collect. Each type of behaviour can fall into the set of ``action type'', which is the  \textbf{Cartesian product} of $A$ and $B$. $A$ = \{train ticket, domestic flight ticket, international flight ticket, vaction item, hotel, search\}, $B$ = \{click, purchase, collect\}. 
We collect user behaviors from all business lines, and formulate each action instance as a 2-tuple $[cityID, action~unit]$ as discussed in Section~\ref{sec:inputlayer}. 
For train ticket and flight ticket, we use the arrival city as the {\em cityID}. 
For vacation items associated with multiple cities, we select the most hot city as the {\em cityID}. 
For hotel and searching, the associated city is used as its {\em cityID}.
These five `action instance' sequences are fused into a global sequence according to corresponding behavior time. 
%Users with abnormal behaviors are removed. 
The statistics of five behavior sequences and the global behavior sequence are illustrated in Table~\ref{tab:behavseq},
where length represents the number of action instances in the sequence.

\begin{table}[htbp]
  \caption{Statistics of five behaviors sequences}\label{tab:behavseq}
  \centering
  \setlength{\tabcolsep}{2mm}{
  \begin{tabular}{|l||c|c|c|}
    \hline
    Sequence& Maximum & Minimum & Average \\
    & length & length & length \\
    \hline
    train& 103& 0& 13.5\\
    flight & 156& 0& 15.9\\
    hotel& 257& 0& 24.9\\
    item& 248& 0& 18.9\\
    search& 66& 0& 6.3\\
    \hline
    \hline
    {\bf global}& 319& 1& 28.6\\
    \hline
  \end{tabular}}
\end{table}

We collect the impression/click data in Fliggy's various recommended scenarios as the label of dataset, the impression and click samples as positive samples, the impression but not click samples as negative samples. Collecting samples exposed by users can effectively reduce noise and ensure users' real interest preferences which is commonly used method of sample collection in real industrial scenes. The statistics of dataset is illustrated in Table~\ref{tab:dataststistic}.

% We collect the impression/click data in Fliggy's various recommended scenarios as the label of dataset, the impression and click samples as positive samples, the impression but not click samples as negative samples. The statistics of dataset is illustrated in Table~\ref{tab:dataststistic}.

\begin{table}[htbp]
  \caption{Statistics of Fliggy's dataset ( M - Million)}\label{tab:dataststistic}
  \centering
  \setlength{\tabcolsep}{2mm}{
  \begin{tabular}{|l||c|c|c|}
    \hline
          & Training & Validation & Testing \\
    \hline
    \# of samples& 26.4M& 3.99M & 4.03M\\
    \# of positive samples & 1.86M& 0.273M&0.274M \\
    \# of negative samples& 24.5M& 3.72M& 3.75M\\
    \# of users& 2.53M& 0.478M& 0.486M\\
    \# of vacation items& 0.254M& 0.185M& 0.173M\\
    \# of destinations & 0.004M& 0.004M& 0.004M\\
    \hline
  \end{tabular}}
\end{table}

\subsubsection{Evaluation Metric}
In order to evaluate the performance of proposed methods, we adopt the Area Under the ROC Curve (AUC) as the evaluation metric. 
AUC is not sensitive to class imbalance and thus is widely used in online recommender systems.
It reflects the probability that a model ranks a randomly chosen positive sample higher than negative samples
and larger AUC represents better performance. 
Exhaustive experiments show that a small improvement in AUC can lead to a significant increase in the online intention prediction accuracy \cite{DBLP:conf/recsys/Cheng0HSCAACCIA16}.

\subsubsection{Competing Models}
%As mentioned above, we evaluate the accuracy of both intention destination prediction and vacation item prediction.
We focus on intention destination prediction in this paper, and intention destinations are important for online intention predictions in recommender systems. 
Thus, we evaluate the accuracy of both intention destination prediction and vacation item prediction.
 
For intention destination prediction, although existing previous intention prediction models focus on recommending vacation items, 
they can be easily adapted to predict intention destination through changing the targets to cityIDs.
Thus, we compare the following models in terms of intention destination prediction:
%
%we compare following models:
%In our experiments, we will compare the following methods for CTR prediction.
\begin{itemize}
%\item \previous{}:
%As far as we know, existing previous intention prediction models focus on recommending vacation items,
%however, they can be easily adapted to predict intention destination through changing the targets
%to cityIDs. And we compete with five previous models:
% \xftrl{} is an eXent Follow-the-Regularized-Leader (FTRL)~\cite{DBLP:conf/colt/McMahanS10,DBLP:journals/jmlr/McMahan11a,DBLP:conf/kdd/McMahanHSYEGNPDGCLWHBK13}  which is implemented by Alibaba group;
% \dnn{} represents Deep Neural Networks implemented over MLP;
 %\wide{}~\cite{DBLP:conf/recsys/Cheng0HSCAACCIA16} is a model proposed by Google;
% \auto{}~\cite{DBLP:conf/cikm/SongS0DX0T19} is an Automatic Feature Interaction neural networks;
% \din{}~\cite{DBLP:conf/kdd/ZhouZSFZMYJLG18} is a Deep Interest Network, which is the latest online serving model in Fliggy. 
\item \auto{}~\cite{DBLP:conf/cikm/SongS0DX0T19} is a state-of-the-art Automatic Feature Interaction neural networks.
\item \din{}~\cite{DBLP:conf/kdd/ZhouZSFZMYJLG18} is a Deep Interest Network, which is the latest online serving model in Fliggy.
\item \mlpit{}: %DIN\&DMSN-MLP combines a DMSN model and DIN to predict CTR. 
In our proposed Deep Multi-Sequence fused neural Networks (i.e., \dmsn{}), five behavior sequences are fused into a global sequence, 
which is used with other auxiliary features to predict the user preference score for a destination. 
\mlp{} implemented the \dmsn{} structure that includes the fully-connected layer and excludes the attention layer.
%\item \rnn{}: Different with \mlpit{}, \rnnit{} uses a Gate Recurrent Unit (GRU) \cite{DBLP:journals/corr/ChungGCB14} layer instead of only use the first full-connected layer. \rnnit{} also excludes the attention layer.
\item \atrnnit{}: As shown in Figure~\ref{fig:TARNN}, attention based recurrent neural network is implemented in \atrnn{}. The RNN layer is implemented by GRU.
%\item \mc{}: \mcit{} uses Multi-grained Convolutional neural network (MC) instead of the GRU layer, and it excludes the attention layer.
\item \atmcit{}: As shown in Figure \ref{fig:ATMC}, attention based multi-grained CNN is implemented in \atmc{}.
\end{itemize}

For vacation item prediction, we merge the proposed intention destination prediction models into existing vacation item recommender systems, and the details will be discussed in Section~\ref{sec:exp_vacationitem}.

\begin{figure*}[!ht]
	\small \center
	\begin{tabular}{@{}c@{ }c@{}}
	\includegraphics[width=0.57\textwidth]{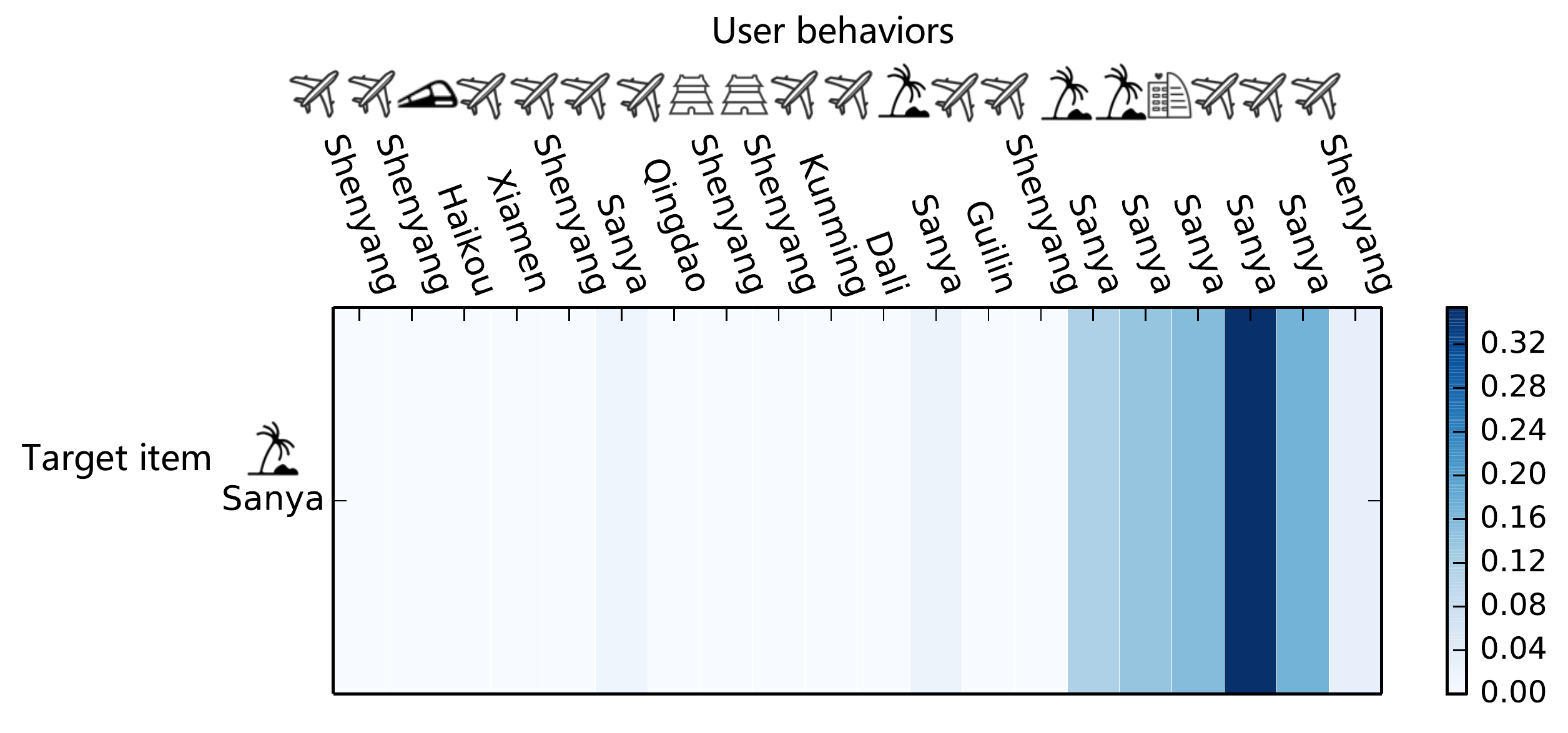} &
	\includegraphics[width=0.4\textwidth]{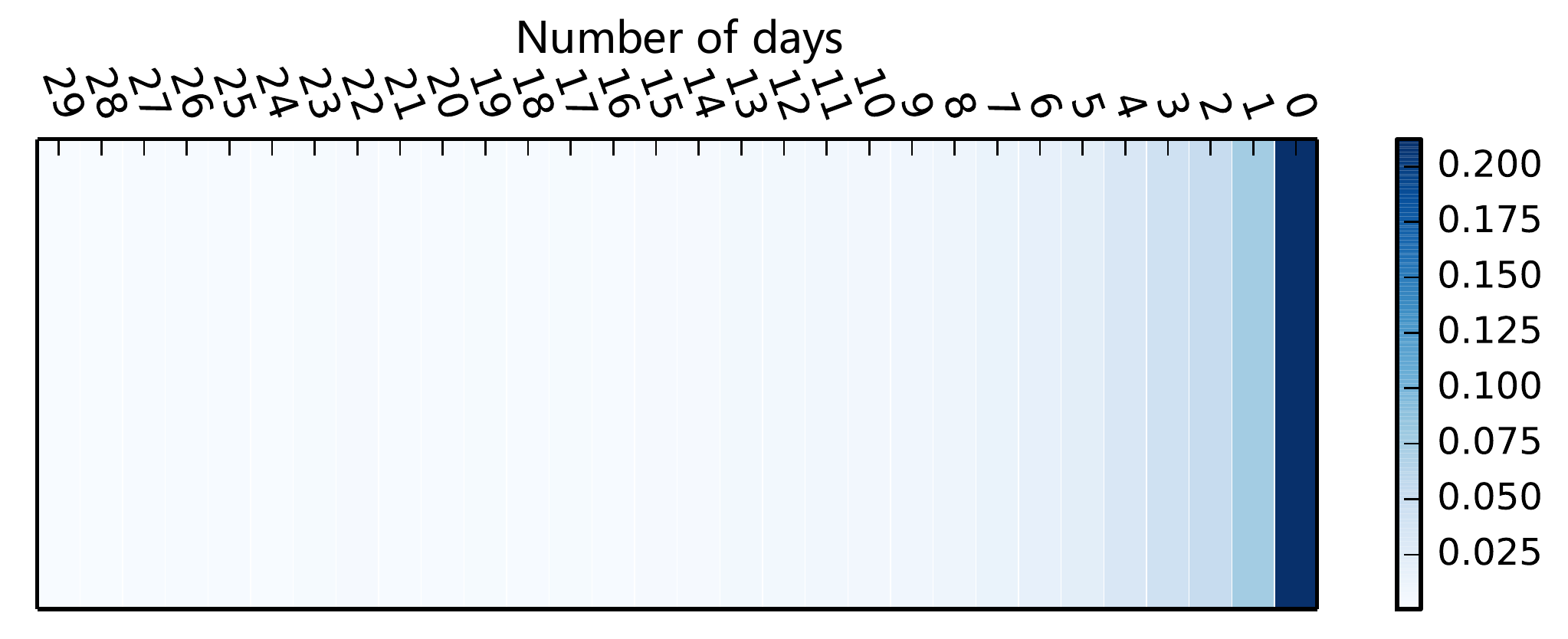} \\
	 (a) [ATRNN] attention weights v.s. behavior locations & (b) [ATRNN]  attention weights v.s. behavior timestamp\\
	\includegraphics[width=0.57\textwidth]{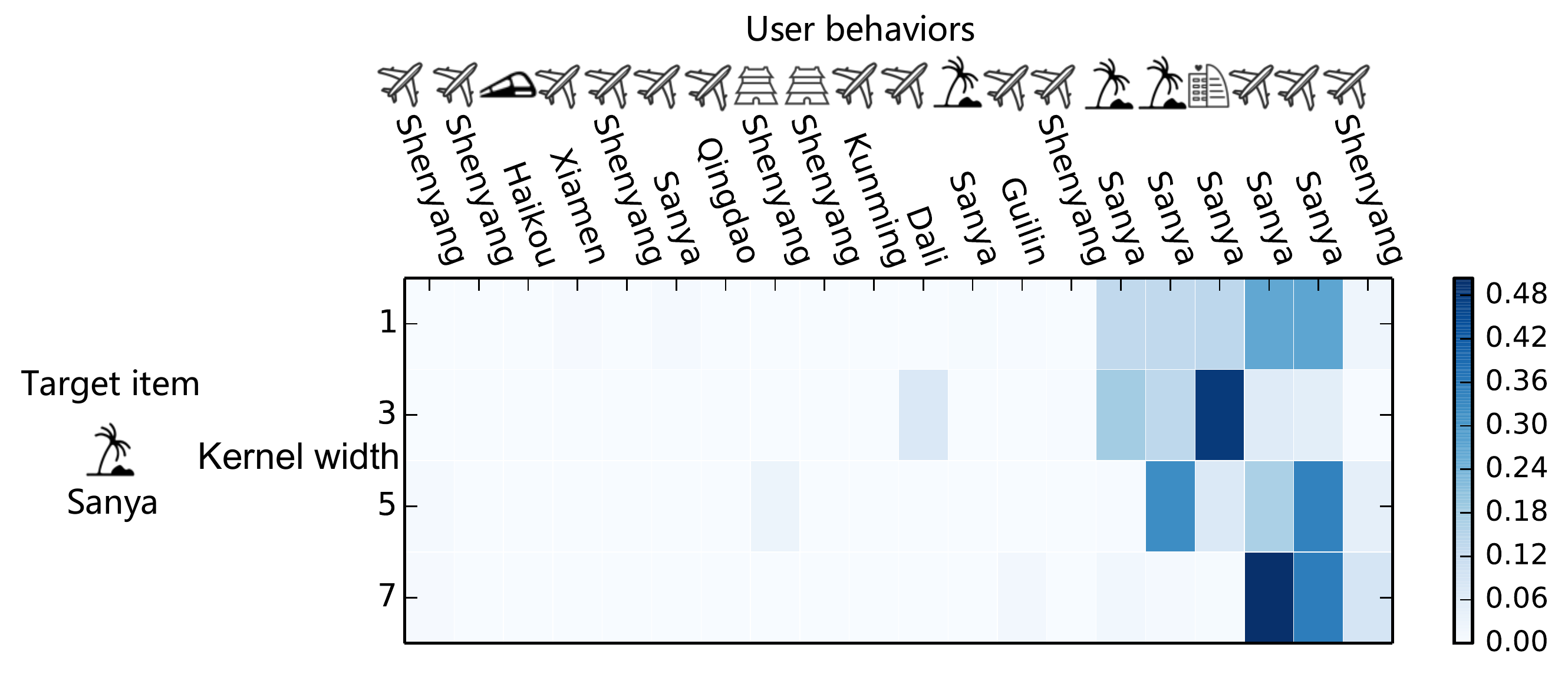} &
	\includegraphics[width=0.4\textwidth]{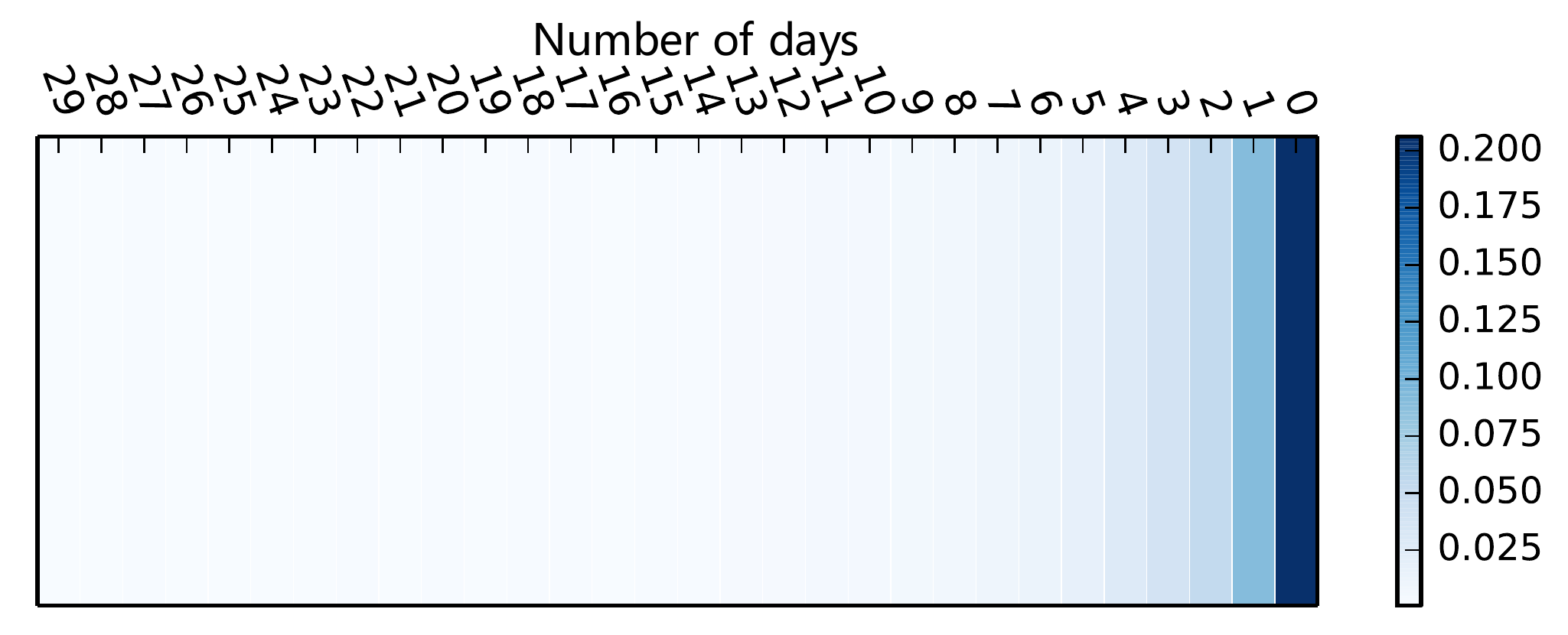} \\
   	 (c) [ATMC] attention weights v.s. behavior locations  & (d) [ATMC] attention weights v.s. behavior timestamp \\
    \end{tabular}
    \caption{Visualized case study for time-aided attention weights of behaviors}\label{fig:visualcase}
\end{figure*}

\subsubsection{Implementation Details}
We set the maximum length of global sequence to 100. 
If the sequence length is less than 100, it will be padded with 0. 
If the sequence length is more than 100, the last 100 `action instance' will be truncated. 
We implement our method using Tensorflow. All hyper-parameters of all models are adjusted by grid-searching on the validation set. 
The dimension of embedding for all models is set to 32. 
The MLP contains four hidden layer, with dimensions 512, 256, 128 and 1. 
The hidden state size of GRU is set to 32. 
We use the Adam with a mini-batch size of 512. 
%In \dmsn{}, we introduce the impression but not click samples as negative samples to our dataset.
%In particular, when implementing the negative sampling mechanism in \dmsn{}, 
%if more than one vacation item impressed at the same time and user clicks the vacation item under $dest_{k}$, the other vacation item under $dest_{k}$ that are not click in this impression can not be used as the negative samples.

\subsection{Visualized Case Study}

%\begin{figure}
%	\small \center
%	\begin{tabular}{@{}c@{}}
%	\includegraphics[width=0.95\columnwidth]{figures/pic_2} \\
%	(a) example \\
  %	\includegraphics[width=0.95\columnwidth]{figures/pic_3} \\
	% (b) time \\
   % \end{tabular}
   % \caption{ATRNN example}\label{fig:rnn_example}
%\end{figure}

To obtain the visualized results, we first train the \atrnnit{} and \atmcit{} models, 
and then choose one user who has acted on a scenic package in Sanya to show the effectiveness of time-aided attention functions.
As discussed in Section~\ref{sec:method},
in \atrnnit{}, a time-aided attention weight vector is calculated for the hidden state according to Equation~\ref{eq:attention1}, 
while one attention weight vector is computed for each of $n_c$ feature maps in \atmcit{}.
The x-axis in Figure~\ref{fig:visualcase}a are user behaviors following the time order from left to right, and 
x-axis in Figure~\ref{fig:visualcase}b illustrate how many days ago the behavior was conducted. 
The color of each bar represents the attention weight of its corresponding behavior, and the weight is calculated by Equation~\ref{eq:attention1}.
The darker the color, the higher the weight. 
As depicted in Figure~\ref{fig:visualcase}a, with our \atrnnit{} model, behaviors in Sanya obtains much higher attention weights than behaviors in other cities, which matches the actual intention destination (i.e., Sanya) of the user.
The result confirms that our proposed attention model can figure out user's actual intention destinations.
Moreover, the time variable used in Equation~\ref{eq:attention1} enables the ATRNN model to capture the effect of time in city prediction.
In detail, as illustrated in Figure~\ref{fig:visualcase}b,
behaviors conducted on the latest day (e.g. searching vacation items and hotels in Sanya, clicking flight tickets from Shenyang) are 
marked to be more important during user intention prediction than behaviors done a month ago.
The result confirms that our proposed attention model can emphasize the time relevance between historical behaviors and users' intentions.

Similar to the results of \atrnnit{},
Figures~\ref{fig:visualcase}c,d depict the visualized results of \atmcit{}. 
As discussed in Section~\ref{subsec:dmsnmodel}, $n_c$
feature maps will be passed to the attention layer in \atmcit{},
where $n_c$ represents the number of different kernels in Multi-Grained CNN Layer.
And for each feature map, one attention vector is calculated.
%After the attention layer, we will obtain $n_{c}$ attention vectors, with one for each feature map.
%We plot those obtained attention weights in Figures~\ref{fig:visualcase}c and~\ref{fig:visualcase}d.
%The user's real intention in this example is scenic packages in Sanya.
And as shown in Figure~\ref{fig:visualcase}c,
we use 4 kernels with width $k=$1, 3, 5, 7,
and for each kernel width, one feature map is passed to the attention layer, and thus
one attention vector is obtained.
Compared with \atrnnit{}, \atmcit{} can assign a higher attention weight to more relevant behaviors, 
for instance the highest weight is 0.48 in Figure~\ref{fig:visualcase}c while the highest is 0.32 in Figure~\ref{fig:visualcase}a.
Moreover, \atmcit{} can explore more relations on product types than \atrnnit{}.
For instance, when $k=1$, behaviors on flight in Sanya are marked as more important, 
while when $k=3$, behaviors on hotels in Sanya are more important. Because in the approaching time of the hotel behavior, the user also acted on vacation item and flight ticket under the same destination. 
That is, with different kernel granularities,
\atmcit{} can capture important features in combination of different product types under the same destination and thus provide more comprehensive and rich data to the MLP layer for accurate prediction.
Besides, similar to \atrnnit{}, \atmcit{} is also sensitive to the conducted time of historical behaviors as shown in Figure~\ref{fig:visualcase}d.

%\begin{figure}
%	\small \center
%	\begin{tabular}{@{}c@{}}
%	\includegraphics[width=0.95\columnwidth]{figures/pic_4} \\
%	(a) example \\
 % 	\includegraphics[width=0.95\columnwidth]{figures/pic_5} \\
%	 (b) time \\
 %   \end{tabular}
 %   \caption{gcnn example}\label{fig:gcnn_example}
%\end{figure}

\subsection{Intention Destination Prediction}
We evaluate the prediction performance over intention destinations.

\begin{figure}[!ht]
	\centering
	\includegraphics[width=0.45\textwidth]{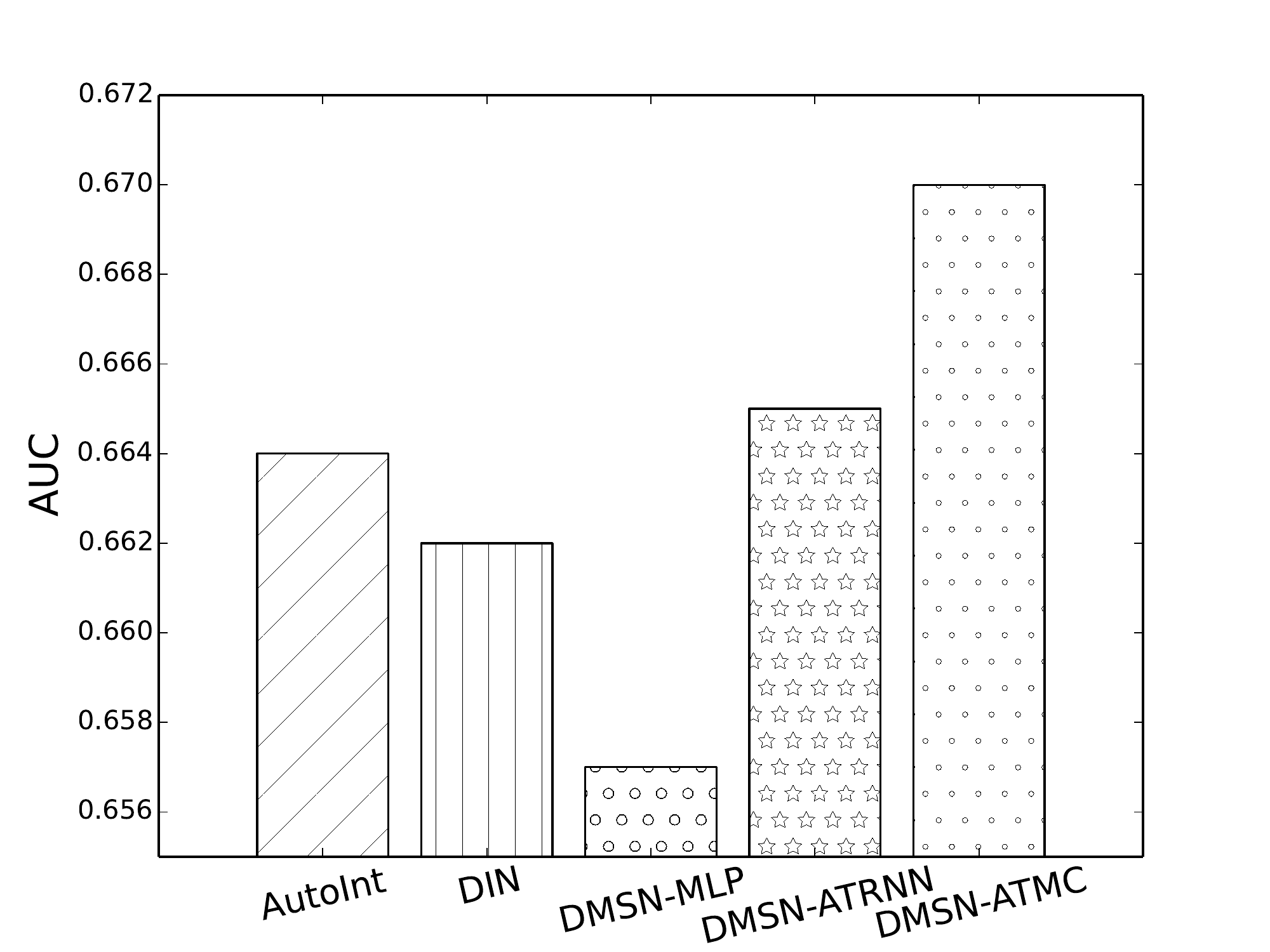}
	\caption{Comparison of proposed models for user's preferred destinations prediction}\label{fig:dmsndest}
\end{figure}

\stitle{Effect of proposed models}
Figure~\ref{fig:dmsndest} 
compares the performance over intention destination prediction.
It is observed that our proposed \atrnnit{} and \atmcit{} perform better than state-of-the-art prediction models.
Besides, \atmcit{} achieves the highest AUC among all competitors. 
The reason is that, differently grained convolution kernels can capture the change pattern 
of users' intention in the long and short term effectively, and can fully capture important features in combination of different product types under the same destination. 

\upsmall

\begin{figure}[!ht]
	\centering
	\includegraphics[width=0.45\textwidth]{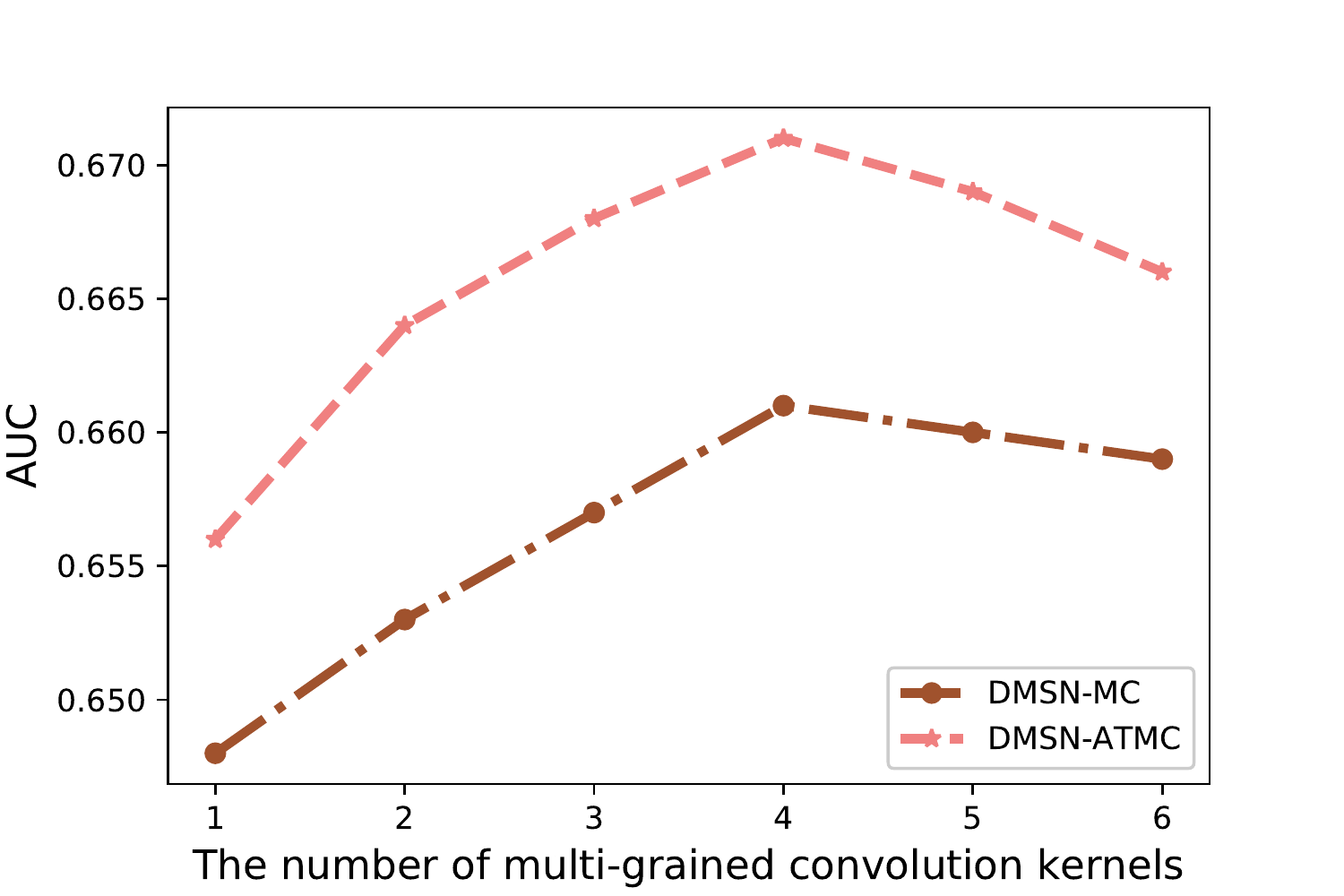}
	\caption{Effect of the number of multi-grained convolution kernels}\label{fig:trend}
\end{figure}

\stitle{Effect of kernel number in \atmcit{}}
The number of multi-grained convolution kernels is an important parameter in multi-grained CNN layer, and thus
we test the performance over different number of kernels as shown in Figure~\ref{fig:trend}.
%Figure~\ref{fig:trend} shows testing AUC with different number of multi-grained convolution kernels. 
We can see that, in the beginning, larger number of kernels induces higher prediction AUC, however,  overfitting occurs when there are too many kernels and thus the prediction AUC drops.
Moreover, \mcit{} represents the model without Attention Layer in \atmcit{},
and from the result, we can see that the attention layer is effective and necessary. 

\begin{figure*}[!ht]
	\small \center
	\begin{tabular}{@{}c@{ }c@{}c@{}}
	\includegraphics[width=0.33\textwidth]{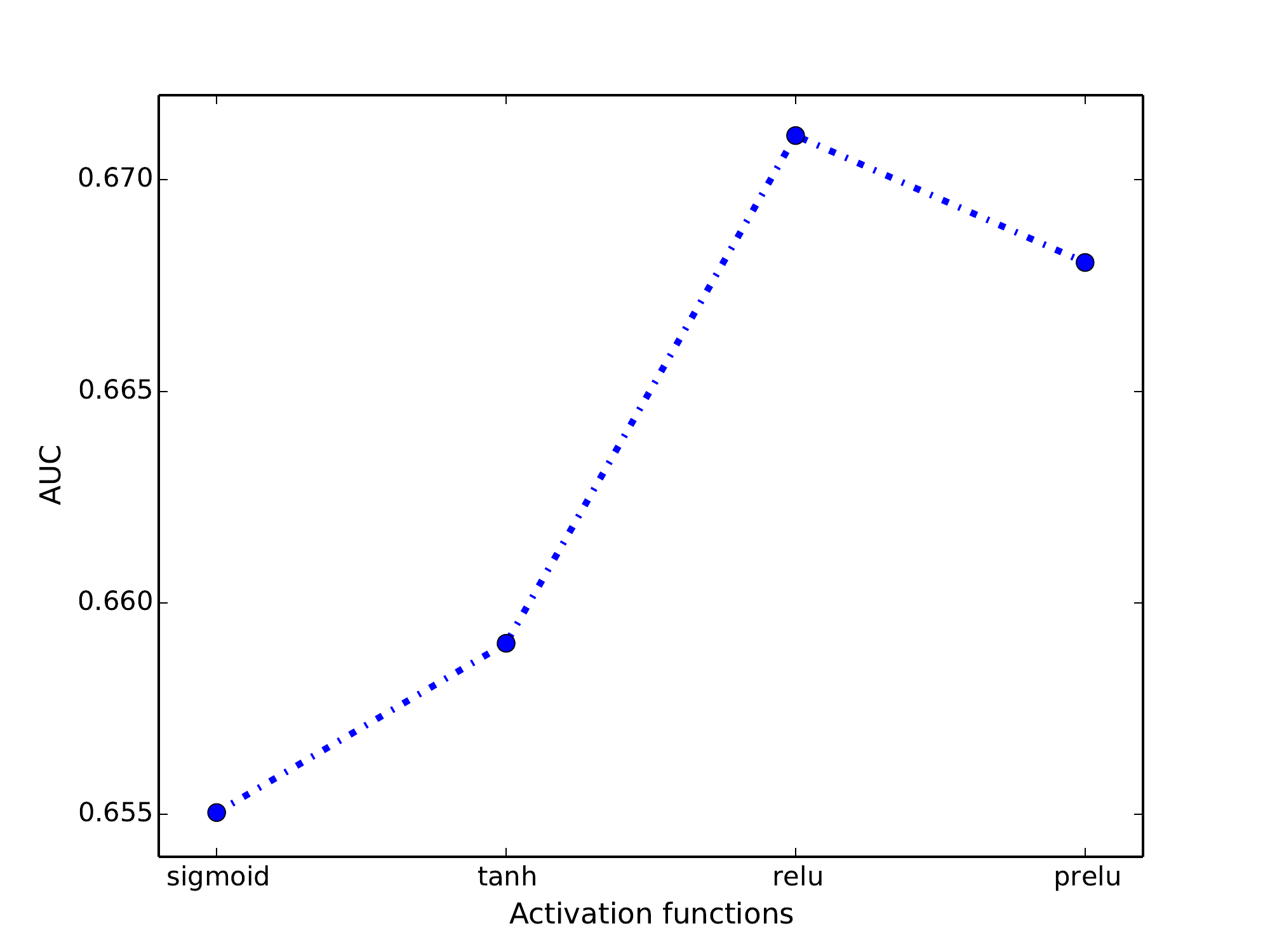} &
	\includegraphics[width=0.33\textwidth]{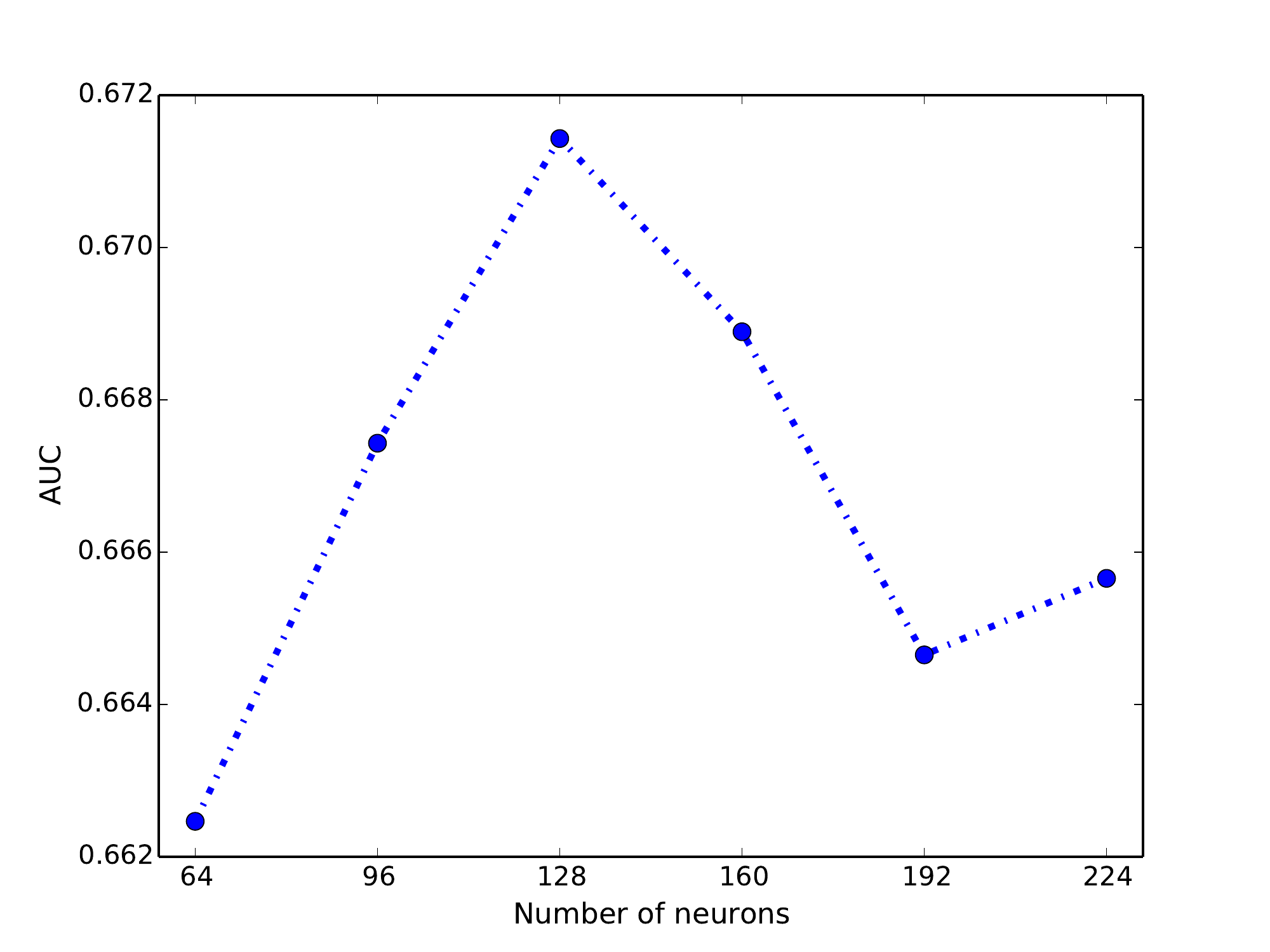} &
	\includegraphics[width=0.33\textwidth]{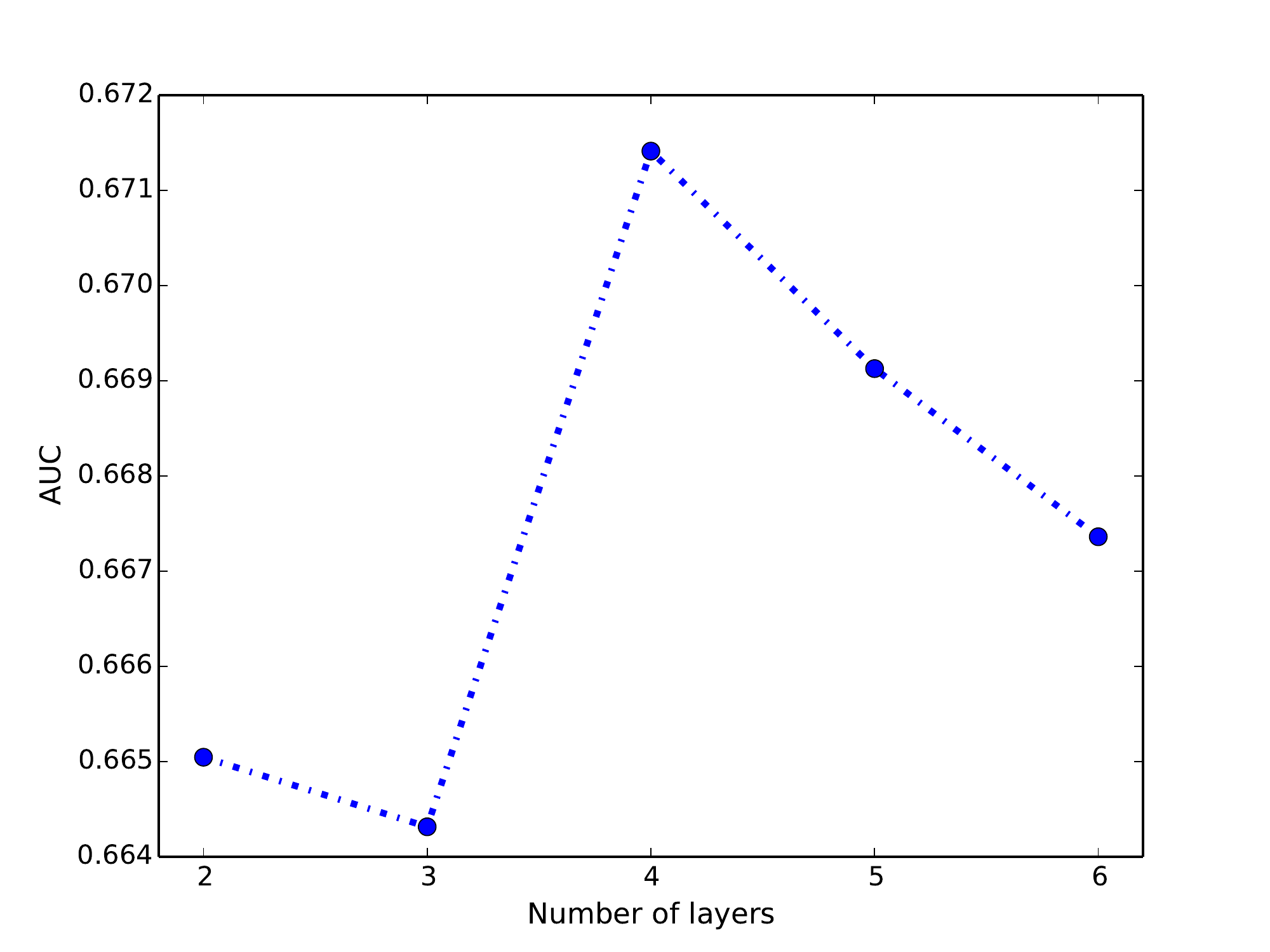} \\
	 (a) Activation functions & (b) Number of neurons & (c) Number of layers
    \end{tabular}
    \caption{Impact of network hyper-parameters on AUC performance}\label{fig:hyper_parameters}
\end{figure*}

\stitle{Effect of hyper-parameters in \atmcit{}}
We study the impact of hyper-parameters of \atmcit{} including the activation functions, the number of neurons in hidden layer and the number of hidden layers. We compare the results of different activation functions in Figure~\ref{fig:hyper_parameters}a and relu is most suitable for \atmcit{}. Moreover, Figure~\ref{fig:hyper_parameters}b demonstrates the impact of number of neurons in hidden layer. When we increase the number of neurons in a fixed setting 4 for the number of hidden layers, 128 is the best setting for \atmcit{}. We can also observe that increasing the number of hidden layers from 2 to 4 can improve the performance. But model performance degrades when the number of hidden layers is set greater than 4 caused by overfitting. 

\stitle{Effect of fusing strategies}

In order to verify the effects of different behavioral sequence fusion strategies, we test whether such fusing is helpful in \atmcit{} by using only one type of sequence, and compare the mentioned two fusing strategies, i.e., {\bf strategy I}: fusing in the very beginning as a global sequence v.s. {\bf strategy II}: fusing after attention layer as five single sequences. In order to align the input format, we use a default behavior sequence of length one to represent the behavior type in which the user has no behavior if we model multiple behavior sequences separately as in {\bf strategy II}. As illustrated in Table~\ref{tab:fusing}, using fused global behavior sequence achieves better performance than using independent feature sequences. And fusing in the very beginning provides the multi-grained CNN layer and attention layer more dense information, thus strategy I outperforms strategy II.
\begin{table}[htbp]
\caption{Effect of feature fusing} \label{tab:fusing}
\centering
\begin{tabular}{|c||c|}
	\hline
	Fusing Strategy & AUC \\
	\hline
	Hotel Single & 0.657 \\
	\hline
	Train Single& 0.66 \\
	\hline
	Flight Single&  0.658 \\
	\hline 
	Item Single& 0.662 \\
	\hline
	Search Single& 0.654 \\
	\hline
	Fusing Strategy I & \bf{0.671} \\
	\hline 
	Fusing Strategy II & 0.665 \\
	\hline
\end{tabular}
\end{table}

\iffalse
\stitle{Effect of number of hidden layers}
To test the robustness of our proposed methods, we vary the number of hidden layers in the MLP Layer in \atmcit{} and compare the  prediction accuracy on users' intention destinations. 
The result in Table~\ref{tab:layernum} depicts that the performance of \atmcit{} model is relatively independent from the number of hidden layers.

\upsmall
\begin{table}[htbp]
  \caption{Effect of number of hidden layers in MLP}\label{tab:layernum}
  \centering
  \begin{tabular}{|c|c|c|c|c|}
     \hline
     num of layers & 5 & 4 & 3 & 2\\
     \hline
     AUC & 0.667 & 0.671 & 0.669 & 0.665\\
     \hline   
  \end{tabular}
\end{table}  
\downsmall
\fi

\begin{figure}[!ht]
	\centering
	\includegraphics[width=0.45\textwidth]{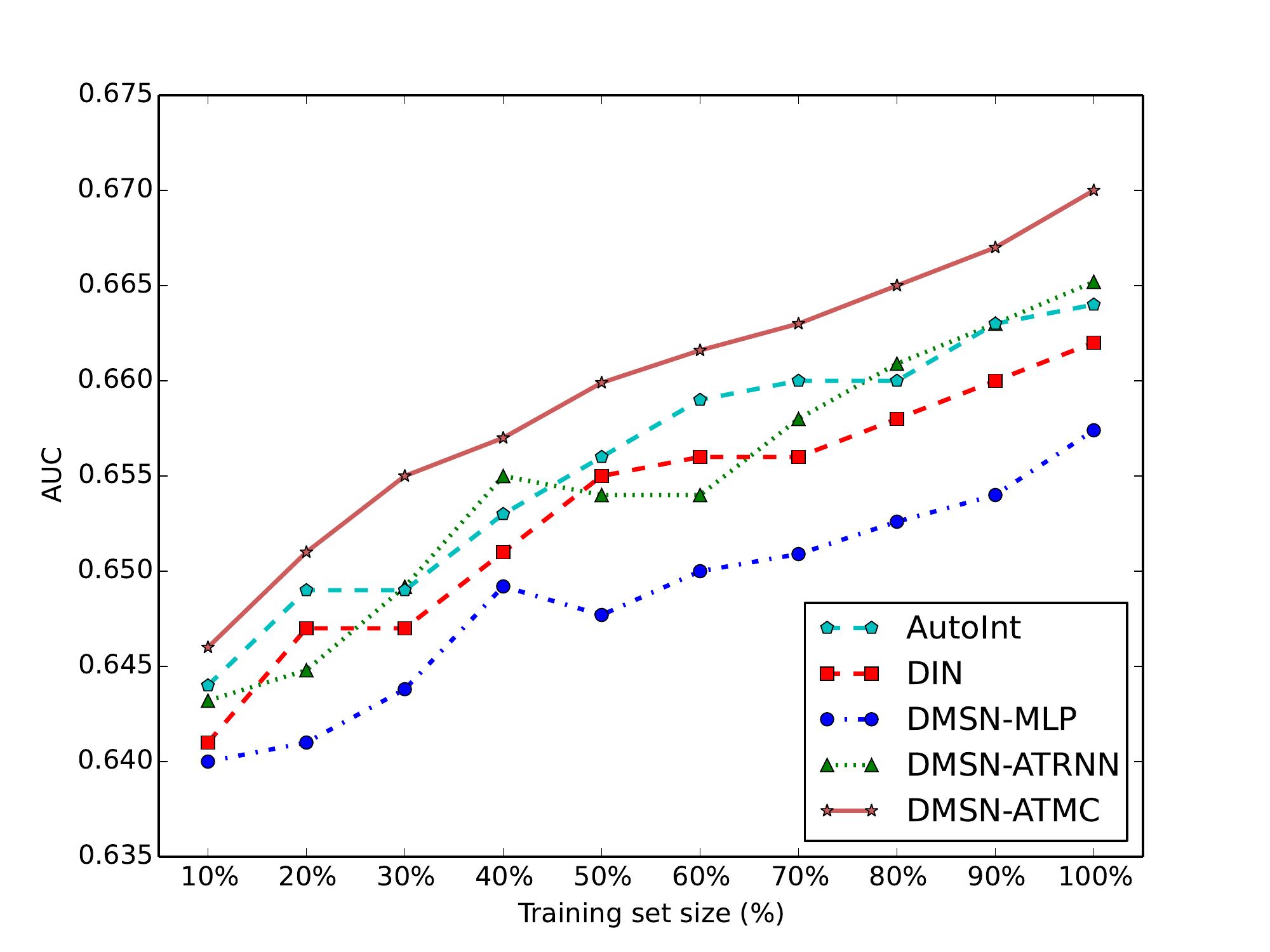}
	\caption{Effect of sampling rates on Fliggy's dataset}\label{fig:auc_samlpe}
\end{figure}

\stitle{Effect of sample size}
Figure~\ref{fig:auc_samlpe} depicts the effect of different size of training sets.  
We compare all competitors in this test and \atmcit{} outperforms other models under all sampling rates. 
With multi-grained CNN layer, \atmcit{} can predict intention destinations mode accurately even with small size of samples.

\subsection{Vacation Item Prediction} \label{sec:exp_vacationitem}
Intention destination predictions are essential for vacation item prediction in online travel platforms. Thus, we also evaluate the effectiveness of our proposed models in terms of vacation item predictions. To do this, we use a two-stage method to merge predicted intention destinations into vacation item predictions. Specifically, we first calculate the preference scores of different destinations, and then estimate the preference scores of vacation items. In detail, the final preference score of $user_{j}$ to vacation item $item_{i}$ is calculated as:

\begin{align}
&p(item_{i}|context_{global}^{j})  \nonumber \\
=& p(dest_{k}|context_{j}^{global})  \times p(item_{i}|context_{j}^{item}) \nonumber \\
=& Score_{Model1} \times Score_{Model2} \nonumber 
\end{align}
where $context_{j}^{global}$ is the global context information of $user_{j}$ in Fliggy,
$p(dest_{k}|context_{j}^{global})$ is the probability that $user_{j}$ will goto destination $dest_{k}$ according to his historical behaviors in $context_{j}^{global}$, and
$p(item_{i}|context_{j}^{item})$ is the probability that $user_{j}$ will click $item_{i}$.

Moreover, both $p(item_{i}|context_{j}^{item})$ and $p(dest_{k}|context_{j}^{global})$ can be obtained by any of the tested five competitor models. 
As \atmcit{} always performs the best in proposed \dmsnit{} models, we exclude \atrnnit{} and \mlpit{} in the following discussions.
 
\begin{figure}[!ht]
	\centering
	\includegraphics[width=0.45\textwidth]{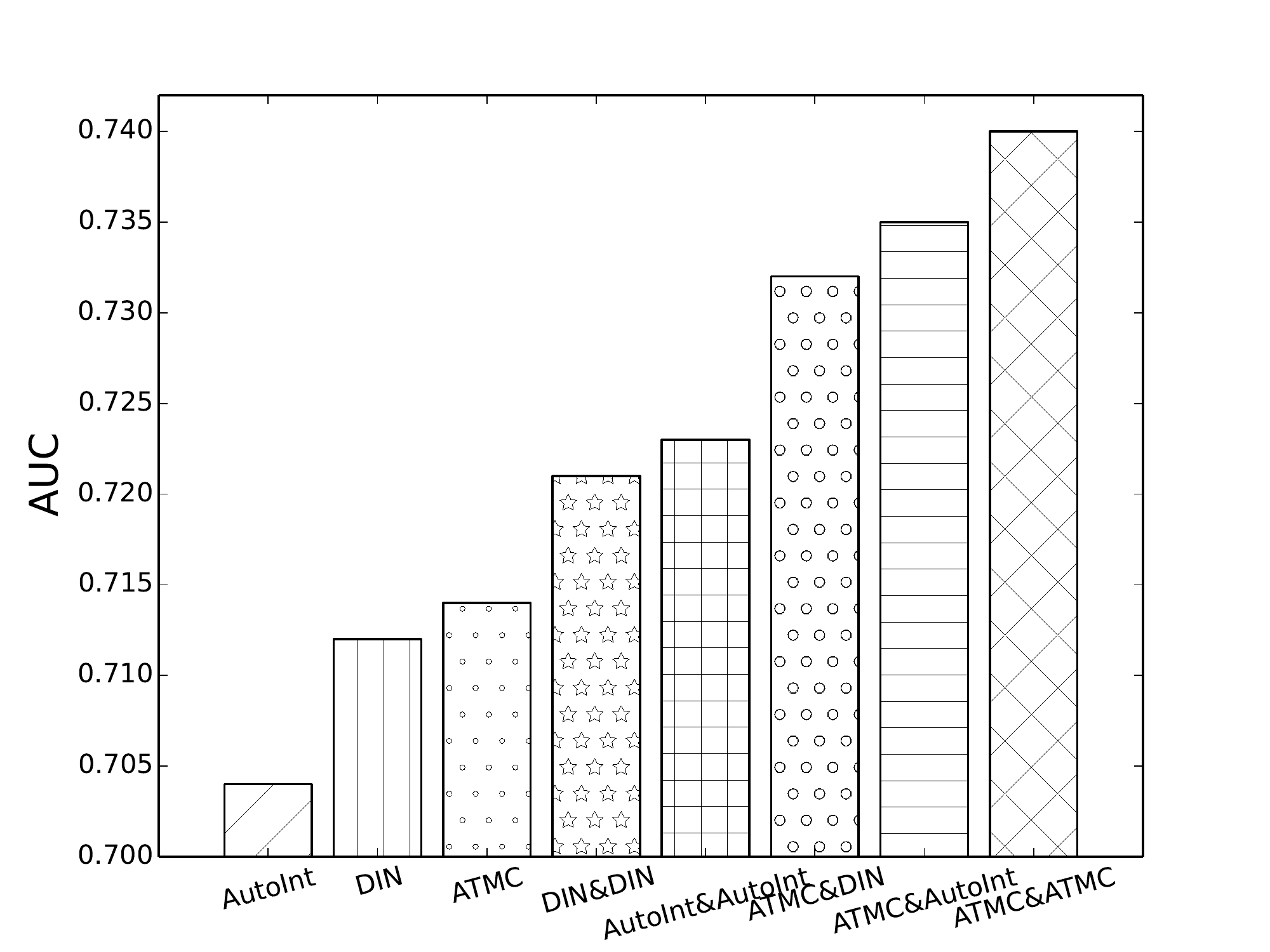}
	\caption{Comparison of different models on Fliggy's Dataset in terms of vacation item prediction}\label{fig:itempredict}
\end{figure}

Figure~\ref{fig:itempredict} illustrates the prediction results over vacation items.
The first three bars using \auto{}, \din{} and \atmcit{} to predict vacation items directly without predicting destinations first.
The remaining bars using 2-stage procedures, and {\em Model1\&Model2} means we use {\em Model1} to predict $p(dest_{k}|context_{j}^{global})$ and {\em Model2} to predict $p(item_{i}|context_{j}^{item})$.
We can see that, 2-stage methods result in higher vacation item prediction accuracy, and this verifies the importance of intention destination predictions.
Moreover, ATMC\&ATMC performs the best, indicating that \dmsnit{} models can effectively improve the ranking results.

\subsection{Online Vacation Item Recommendation}
We deploy our model on Fliggy's online recommender platform, and conduct experiments on standard A/B testing environment. 
We use the real-time click-through rate (CTR) as the metric to evaluate the performance of our proposed ATMC$\&$ATMC model.
We ran the experiment for one week on each business line in Fliggy.
As shown in Table~\ref{Table5},  our proposed ATMC\&ATMC achieves significant CTR improvement on four business lines, and achieves the highest CTR improvement on train ticket business line.

\begin{table}[htbp]
  \caption{Average relative improvement results of CTR}\label{Table5}
  \centering
  \setlength{\tabcolsep}{2mm}{
  \begin{tabular}{|l||c|}
    \hline
     Business line & CTR improvement \\
    \hline
    Train & +3.4\%\\
    \hline
    Flight& +2.5\%\\
    \hline
    Vacation items& +1.4\%\\
    \hline
    Hotel& +1.4\%\\
    \hline
  \end{tabular}}
\end{table}

\section{Related Work} \label{sec:related}
This paper studies how to provide satisfactory travel plannings for users,
especially we focus on predicting users' real intention destinations in traveling scenario.
%This paper studies the click-through rate prediction problem in traveling scenario.
Intention prediction has attracted attentions from both industry and academic communities.
As many internet companies utilize click-through rates to predict users' preferences, various prediction systems have been developed for online intention predictions~\cite{DBLP:conf/recsys/Cheng0HSCAACCIA16, paper1cite10, paper1cite15, DBLP:conf/kdd/McMahanHSYEGNPDGCLWHBK13, paper1cite29, DBLP:conf/kdd/ZhouZSFZMYJLG18}.
%deleted before
Wide$\&$Deep learning system~\cite{DBLP:conf/recsys/Cheng0HSCAACCIA16} proposed by Google combines the advantages of both the linear shallow model and deep models for recommender systems.
Wide$\&$Deep model achieves remarkable performance in APP recommendation.
DIN~\cite{DBLP:conf/kdd/ZhouZSFZMYJLG18} model proposed by Alibaba group is designed for online advertising.
Different from other click-through rate models which compress user features into a fixed-length representation vector, 
DIN designs a local activation unit to adaptively learn the representation of user interests from historical behaviors with respect to a certain ad.
%deleted up
Besides industry communities, intention prediction is also well studied in academic communities~\cite{paper1cite31,paper1cite24}. 
A context aware click-through rate prediction method ~\cite{paper1cite31} is proposed with factorized three-way
$\langle user, ad, context \rangle$
%{\color{blue}{<user, ad, context>(尖括号显示问题)}} 
tensor.
Hierarchical importance-aware factorization machine~\cite{paper1cite24} is developed to model dynamic impacts of ads prediction.
%deleted up
%
The structure of intention prediction model has evolved from shallow to deep.
And the number of samples and the dimension of features used in intention prediction model have become larger and larger.
Thus, more and more model structures are designed to improve the feature extraction performance~\cite{paper2cite2,paper2cite8,paper1cite26,paper1cite13,paper2cite20,DBLP:conf/recsys/Cheng0HSCAACCIA16,paper2cite3,DBLP:conf/kdd/ZhouZSFZMYJLG18}.
%deleted down
The widely used base click-through rate prediction model is a combination of embedding layer (for learning the dense representation of sparse features) and MLP (for learning the combination relations of features automatically).
The embedding method is pioneered discussed in NNLM~\cite{paper2cite2}, which learns distributed representation for each word to avoid curse of dimension in language modeling.
Dense representations of features after embedding layer are interacted using specially designed transformation functions for target fitting 
LS-PLM~\cite{paper2cite8} and FM~\cite{paper1cite26} models are a class of networks with one hidden layer, which captures the first- and second-order feature interactions in recommender systems
%deleted up
%
%deleted down
Variants of factorization machines include Field-aware Factorization Machines (FFM)~\cite{paper1cite16},GBFM~\cite{paper1cite7} and AFM\cite{paper1cite40}.
FFM models fine-grained interactions between features from different fields, GBFM and AFM considered the importance of different second-order feature interactions.
All these base models only focus on low-order feature interactions.
In terms of high-order feature interactions, NFM\cite{paper1cite13} stacked deep neural networks on top of the output of the second-order feature interactions to model higher-order features.
Moreover, Deep Crossing~\cite{paper2cite20}, Wide$\&$Deep Learning~\cite{DBLP:conf/recsys/Cheng0HSCAACCIA16} and YouTube Recommendation click-through rate model~\cite{paper2cite3} 
extend
LS-PLM and FM by replacing the transformation function with complex MLP network, which enhances the model capability greatly.
PNN~\cite{paper2cite5} tries to capture high-order feature interactions by involving a product layer after embedding layer. 
DeepFM~\cite{DBLP:conf/ijcai/GuoTYLH17} imposes a factorization machines as "wide" module in Wide$\&$Deep with no need of feature engineering. 
Deep$\&$Cross~\cite{paper1cite38} and xDeepFM~\cite{paper1cite19} took outer product of features at the bit- and vector-wise level respectively to learn feature interactions. 
All above approaches are not trivial to explain which combinations are useful, thus the work~\cite{DBLP:conf/cikm/SongS0DX0T19} explicitly models feature interactions with attention mechanism in an end-to-end manner, and probe the learned feature combinations via visualization.
For applications with rich user behaviors, features are often contained with variable-length list of ids, 
e.g., searched terms or watched videos in YouTube recommender systems.
Above works often transform corresponding list of embedding vectors into a fixed-length vector, which may cause loss of information.
DIN~\cite{DBLP:conf/kdd/ZhouZSFZMYJLG18} tackles it by adaptively learning the representation vector w.r.t. given ad, improving the expressive ability of model.
%deleted up
The proposed DMSN models in this paper use the latest technique in the literature of deep learning which is attention mechanism~\cite{paper1cite2}.
Aided with attention-based neural networks,
the DMSN models can get an expected annotation and 
focuses only on information relevant to the generation of next target word.
%deleted down
Attention is first proposed in the context of neural machine translation~\cite{paper1cite2} and has been proved effective in a variety of tasks such as neural machine translation and recommender systems. 
Neural Machine Translation (NMT)~\cite{paper1cite2} takes a weighted sum of all the annotations to get an expected annotation and focuses only on information relevant to the generation of next target word.
DeepIntent~\cite{paper2cite25} applies attention in the context of search advertising and it uses RNN to model text to help paying attention on the key words in each query.
DIN~\cite{DBLP:conf/kdd/ZhouZSFZMYJLG18} designs a local activation unit to soft-search for relevant user behaviors and takes a weighted sum pooling to obtain the adaptive representation of user interests with respect to a given ad.
%deleted up
%
Compared with existing works,
this paper exploits the characteristics of online travel platforms and 
propose to integrate attention-based neural networks into multi-sequence fused frameworks for 
users' intention destination prediction.
Moreover, there exist related literatures studying destination predictions~\cite{dest_1, dest_2, dest_3, dest_4}.
~\cite{dest_1} utilizes a hidden Markov model for predicting driver destinations and routes.
~\cite{dest_2} proposes a hybrid model to predict future gathering events through trajectory destination prediction.
~\cite{dest_3} studies a novel data embedding method and ensemble learning method to provide accurate and timely destination predictions of taxis.
~\cite{dest_4} utilizes trust-enhanced collaborative filtering to predict users' intention POIs according to their historical trajectory behaviors.
However, none of existing methods consider to use users' multi-category behaviors for online travelling destination prediction.

\up
\section{Conclusion} \label{sec:conclude}
In this paper, we focus on predicting users' intention destinations during travel plannings in online travel platforms.
Accurate predicted destinations can significantly improve the vacation item recommendation performance. 
%model the process of CTR prediction in online travel platforms as 
%$context\to destination\to vacation~items$
In detail, we propose two models to predict users' real intention destinations.
Both models follow the Deep Multi-Squences fused neural Networks (DMSN) architecture, 
which fuses different categories of user behavior sequences into a global cityID sequence and use the global sequence to 
predict destination preferences. 
To emphasize users' preference for different destinations during different periods, 
we intersect the DMSN model with attention-based recurrent neural networks (ATRNN) and 
attention-based multi-grained convolutional neural networks(ATMC).
Experimental results on real data from Fliggy's offline log and online A/B testing illustrate the effectiveness of our DMSN models.

\ifCLASSOPTIONcaptionsoff
  \newpage
\fi



\bibliographystyle{IEEEtran}
\bibliography{ref}

\begin{thebibliography}{10}
\providecommand{\url}[1]{#1}
\csname url@samestyle\endcsname
\providecommand{\newblock}{\relax}
\providecommand{\bibinfo}[2]{#2}
\providecommand{\BIBentrySTDinterwordspacing}{\spaceskip=0pt\relax}
\providecommand{\BIBentryALTinterwordstretchfactor}{4}
\providecommand{\BIBentryALTinterwordspacing}{\spaceskip=\fontdimen2\font plus
\BIBentryALTinterwordstretchfactor\fontdimen3\font minus
  \fontdimen4\font\relax}
\providecommand{\BIBforeignlanguage}[2]{{%
\expandafter\ifx\csname l@#1\endcsname\relax
\typeout{** WARNING: IEEEtran.bst: No hyphenation pattern has been}%
\typeout{** loaded for the language `#1'. Using the pattern for}%
\typeout{** the default language instead.}%
\else
\language=\csname l@#1\endcsname
\fi
#2}}
\providecommand{\BIBdecl}{\relax}
\BIBdecl

\bibitem{DBLP:conf/kdd/ZhuLZLHLG18}
\BIBentryALTinterwordspacing
H.~Zhu, X.~Li, P.~Zhang, G.~Li, J.~He, H.~Li, and K.~Gai, ``Learning tree-based
  deep model for recommender systems,'' in \emph{Proceedings of the 24th {ACM}
  {SIGKDD} International Conference on Knowledge Discovery {\&} Data Mining},
  2018, pp. 1079--1088. [Online]. Available:
  \url{https://doi.org/10.1145/3219819.3219826}
\BIBentrySTDinterwordspacing

\bibitem{DBLP:journals/computer/KorenBV09}
\BIBentryALTinterwordspacing
Y.~Koren, R.~M. Bell, and C.~Volinsky, ``Matrix factorization techniques for
  recommender systems,'' \emph{{IEEE} Computer}, vol.~42, no.~8, pp. 30--37,
  2009. [Online]. Available: \url{https://doi.org/10.1109/MC.2009.263}
\BIBentrySTDinterwordspacing

\bibitem{DBLP:conf/ijcai/GuoTYLH17}
\BIBentryALTinterwordspacing
H.~Guo, R.~Tang, Y.~Ye, Z.~Li, and X.~He, ``Deepfm: {A} factorization-machine
  based neural network for {CTR} prediction,'' in \emph{Proceedings of the
  Twenty-Sixth International Joint Conference on Artificial Intelligence},
  2017, pp. 1725--1731. [Online]. Available:
  \url{https://doi.org/10.24963/ijcai.2017/239}
\BIBentrySTDinterwordspacing

\bibitem{DBLP:conf/recsys/Cheng0HSCAACCIA16}
\BIBentryALTinterwordspacing
H.~Cheng, L.~Koc, J.~Harmsen, T.~Shaked, T.~Chandra, H.~Aradhye, G.~Anderson,
  G.~Corrado, W.~Chai, M.~Ispir, R.~Anil, Z.~Haque, L.~Hong, V.~Jain, X.~Liu,
  and H.~Shah, ``Wide {\&} deep learning for recommender systems,'' in
  \emph{Proceedings of the 1st Workshop on Deep Learning for Recommender
  Systems}, 2016, pp. 7--10. [Online]. Available:
  \url{https://doi.org/10.1145/2988450.2988454}
\BIBentrySTDinterwordspacing

\bibitem{DBLP:conf/kdd/ZhouZSFZMYJLG18}
G.~Zhou, X.~Zhu, C.~Song, Y.~Fan, H.~Zhu, X.~Ma, Y.~Yan, J.~Jin, H.~Li, and
  K.~Gai, ``Deep interest network for click-through rate prediction,'' in
  \emph{Proceedings of the 24th {ACM} {SIGKDD} International Conference on
  Knowledge Discovery {\&} Data Mining}, 2018, pp. 1059--1068.

\bibitem{DBLP:conf/cikm/SongS0DX0T19}
\BIBentryALTinterwordspacing
W.~Song, C.~Shi, Z.~Xiao, Z.~Duan, Y.~Xu, M.~Zhang, and J.~Tang, ``Autoint:
  Automatic feature interaction learning via self-attentive neural networks,''
  in \emph{Proceedings of the 28th {ACM} International Conference on
  Information and Knowledge Management, {CIKM} 2019, Beijing, China, November
  3-7, 2019}, 2019, pp. 1161--1170. [Online]. Available:
  \url{https://doi.org/10.1145/3357384.3357925}
\BIBentrySTDinterwordspacing

\bibitem{DBLP:conf/kdd/OuyangZLZXLD19}
W.~Ouyang, X.~Zhang, L.~Li, H.~Zou, X.~Xing, Z.~Liu, and Y.~Du, ``Deep
  spatio-temporal neural networks for click-through rate prediction,'' in
  \emph{Proceedings of the 25th {ACM} {SIGKDD} International Conference on
  Knowledge Discovery {\&} Data Mining}, 2019, pp. 2078--2086.

\bibitem{DBLP:conf/sigir/MaZHWHZG18}
\BIBentryALTinterwordspacing
X.~Ma, L.~Zhao, G.~Huang, Z.~Wang, Z.~Hu, X.~Zhu, and K.~Gai, ``Entire space
  multi-task model: An effective approach for estimating post-click conversion
  rate,'' in \emph{The 41st International {ACM} {SIGIR} Conference on Research
  {\&} Development in Information Retrieval}, 2018, pp. 1137--1140. [Online].
  Available: \url{https://doi.org/10.1145/3209978.3210104}
\BIBentrySTDinterwordspacing

\bibitem{DBLP:conf/aaai/WenZLYH19}
\BIBentryALTinterwordspacing
H.~Wen, J.~Zhang, Q.~Lin, K.~Yang, and P.~Huang, ``Multi-level deep cascade
  trees for conversion rate prediction in recommendation system,'' in \emph{The
  Thirty-Third {AAAI} Conference on Artificial Intelligence}, 2019, pp.
  338--345. [Online]. Available:
  \url{https://doi.org/10.1609/aaai.v33i01.3301338}
\BIBentrySTDinterwordspacing

\bibitem{DBLP:conf/sigir/Shan0S18}
\BIBentryALTinterwordspacing
L.~Shan, L.~Lin, and C.~Sun, ``Combined regression and tripletwise learning for
  conversion rate prediction in real-time bidding advertising,'' in \emph{The
  41st International {ACM} {SIGIR} Conference on Research {\&} Development in
  Information Retrieval}, 2018, pp. 115--123. [Online]. Available:
  \url{https://doi.org/10.1145/3209978.3210062}
\BIBentrySTDinterwordspacing

\bibitem{paper2cite20}
\BIBentryALTinterwordspacing
Y.~Shan, T.~R. Hoens, J.~Jiao, H.~Wang, D.~Yu, and J.~C. Mao, ``Deep crossing:
  Web-scale modeling without manually crafted combinatorial features,'' in
  \emph{Proceedings of the 22nd {ACM} {SIGKDD} International Conference on
  Knowledge Discovery and Data Mining, San Francisco, CA, USA, August 13-17,
  2016}, B.~Krishnapuram, M.~Shah, A.~J. Smola, C.~C. Aggarwal, D.~Shen, and
  R.~Rastogi, Eds.\hskip 1em plus 0.5em minus 0.4em\relax {ACM}, 2016, pp.
  255--262. [Online]. Available: \url{https://doi.org/10.1145/2939672.2939704}
\BIBentrySTDinterwordspacing

\bibitem{DBLP:conf/nips/MikolovSCCD13}
\BIBentryALTinterwordspacing
T.~Mikolov, I.~Sutskever, K.~Chen, G.~S. Corrado, and J.~Dean, ``Distributed
  representations of words and phrases and their compositionality,'' in
  \emph{Advances in Neural Information Processing Systems 26: 27th Annual
  Conference on Neural Information Processing Systems 2013}, 2013, pp.
  3111--3119. [Online]. Available:
  \url{http://papers.nips.cc/paper/5021-distributed-representations-of-words-and-phrases-and-their-compositionality}
\BIBentrySTDinterwordspacing

\bibitem{DBLP:conf/iclr/KimDHR17}
\BIBentryALTinterwordspacing
Y.~Kim, C.~Denton, L.~Hoang, and A.~M. Rush, ``Structured attention networks,''
  in \emph{5th International Conference on Learning Representations}, 2017.
  [Online]. Available: \url{https://openreview.net/forum?id=HkE0Nvqlg}
\BIBentrySTDinterwordspacing

\bibitem{DBLP:conf/nips/VaswaniSPUJGKP17}
\BIBentryALTinterwordspacing
A.~Vaswani, N.~Shazeer, N.~Parmar, J.~Uszkoreit, L.~Jones, A.~N. Gomez,
  L.~Kaiser, and I.~Polosukhin, ``Attention is all you need,'' in
  \emph{Advances in Neural Information Processing Systems 30: Annual Conference
  on Neural Information Processing Systems}, 2017, pp. 5998--6008. [Online].
  Available: \url{http://papers.nips.cc/paper/7181-attention-is-all-you-need}
\BIBentrySTDinterwordspacing

\bibitem{DBLP:conf/ijcai/ZhouF17}
\BIBentryALTinterwordspacing
Z.~Zhou and J.~Feng, ``Deep forest: Towards an alternative to deep neural
  networks,'' in \emph{Proceedings of the Twenty-Sixth International Joint
  Conference on Artificial Intelligence}, 2017, pp. 3553--3559. [Online].
  Available: \url{https://doi.org/10.24963/ijcai.2017/497}
\BIBentrySTDinterwordspacing

\bibitem{paper1cite10}
\BIBentryALTinterwordspacing
T.~Graepel, J.~Q. Candela, T.~Borchert, and R.~Herbrich, ``Web-scale bayesian
  click-through rate prediction for sponsored search advertising in microsoft's
  bing search engine,'' in \emph{Proceedings of the 27th International
  Conference on Machine Learning (ICML-10), June 21-24, 2010, Haifa, Israel},
  J.~F{\"{u}}rnkranz and T.~Joachims, Eds.\hskip 1em plus 0.5em minus
  0.4em\relax Omnipress, 2010, pp. 13--20. [Online]. Available:
  \url{https://icml.cc/Conferences/2010/papers/901.pdf}
\BIBentrySTDinterwordspacing

\bibitem{paper1cite15}
\BIBentryALTinterwordspacing
X.~He, J.~Pan, O.~Jin, T.~Xu, B.~Liu, T.~Xu, Y.~Shi, A.~Atallah, R.~Herbrich,
  S.~Bowers, and J.~Q. Candela, ``Practical lessons from predicting clicks on
  ads at facebook,'' in \emph{Proceedings of the Eighth International Workshop
  on Data Mining for Online Advertising, {ADKDD} 2014, August 24, 2014, New
  York City, New York, {USA}}, E.~Saka, D.~Shen, K.~Lee, and Y.~Li, Eds.\hskip
  1em plus 0.5em minus 0.4em\relax {ACM}, 2014, pp. 5:1--5:9. [Online].
  Available: \url{https://doi.org/10.1145/2648584.2648589}
\BIBentrySTDinterwordspacing

\bibitem{DBLP:conf/kdd/McMahanHSYEGNPDGCLWHBK13}
\BIBentryALTinterwordspacing
H.~B. McMahan, G.~Holt, D.~Sculley, M.~Young, D.~Ebner, J.~Grady, L.~Nie,
  T.~Phillips, E.~Davydov, D.~Golovin, S.~Chikkerur, D.~Liu, M.~Wattenberg,
  A.~M. Hrafnkelsson, T.~Boulos, and J.~Kubica, ``Ad click prediction: a view
  from the trenches,'' in \emph{The 19th {ACM} {SIGKDD} International
  Conference on Knowledge Discovery and Data Mining}, 2013, pp. 1222--1230.
  [Online]. Available: \url{https://doi.org/10.1145/2487575.2488200}
\BIBentrySTDinterwordspacing

\bibitem{paper1cite29}
\BIBentryALTinterwordspacing
M.~Richardson, E.~Dominowska, and R.~Ragno, ``Predicting clicks: estimating the
  click-through rate for new ads,'' in \emph{Proceedings of the 16th
  International Conference on World Wide Web, {WWW} 2007, Banff, Alberta,
  Canada, May 8-12, 2007}, C.~L. Williamson, M.~E. Zurko, P.~F.
  Patel{-}Schneider, and P.~J. Shenoy, Eds.\hskip 1em plus 0.5em minus
  0.4em\relax {ACM}, 2007, pp. 521--530. [Online]. Available:
  \url{https://doi.org/10.1145/1242572.1242643}
\BIBentrySTDinterwordspacing

\bibitem{paper1cite31}
\BIBentryALTinterwordspacing
L.~Shan, L.~Lin, C.~Sun, and X.~Wang, ``Predicting ad click-through rates via
  feature-based fully coupled interaction tensor factorization,''
  \emph{Electron. Commer. Res. Appl.}, vol.~16, pp. 30--42, 2016. [Online].
  Available: \url{https://doi.org/10.1016/j.elerap.2016.01.004}
\BIBentrySTDinterwordspacing

\bibitem{paper1cite24}
\BIBentryALTinterwordspacing
R.~J. Oentaryo, E.~Lim, J.~Low, D.~Lo, and M.~Finegold, ``Predicting response
  in mobile advertising with hierarchical importance-aware factorization
  machine,'' in \emph{Seventh {ACM} International Conference on Web Search and
  Data Mining, {WSDM} 2014, New York, NY, USA, February 24-28, 2014},
  B.~Carterette, F.~Diaz, C.~Castillo, and D.~Metzler, Eds.\hskip 1em plus
  0.5em minus 0.4em\relax {ACM}, 2014, pp. 123--132. [Online]. Available:
  \url{https://doi.org/10.1145/2556195.2556240}
\BIBentrySTDinterwordspacing

\bibitem{paper2cite2}
\BIBentryALTinterwordspacing
Y.~Bengio, R.~Ducharme, P.~Vincent, and C.~Janvin, ``A neural probabilistic
  language model,'' \emph{J. Mach. Learn. Res.}, vol.~3, pp. 1137--1155, 2003.
  [Online]. Available: \url{http://jmlr.org/papers/v3/bengio03a.html}
\BIBentrySTDinterwordspacing

\bibitem{paper2cite8}
\BIBentryALTinterwordspacing
K.~Gai, X.~Zhu, H.~Li, K.~Liu, and Z.~Wang, ``Learning piece-wise linear models
  from large scale data for ad click prediction,'' \emph{CoRR}, vol.
  abs/1704.05194, 2017. [Online]. Available:
  \url{http://arxiv.org/abs/1704.05194}
\BIBentrySTDinterwordspacing

\bibitem{paper1cite26}
\BIBentryALTinterwordspacing
S.~Rendle, ``Factorization machines,'' in \emph{{ICDM} 2010, The 10th {IEEE}
  International Conference on Data Mining, Sydney, Australia, 14-17 December
  2010}, G.~I. Webb, B.~Liu, C.~Zhang, D.~Gunopulos, and X.~Wu, Eds.\hskip 1em
  plus 0.5em minus 0.4em\relax {IEEE} Computer Society, 2010, pp. 995--1000.
  [Online]. Available: \url{https://doi.org/10.1109/ICDM.2010.127}
\BIBentrySTDinterwordspacing

\bibitem{paper1cite13}
\BIBentryALTinterwordspacing
X.~He and T.~Chua, ``Neural factorization machines for sparse predictive
  analytics,'' in \emph{Proceedings of the 40th International {ACM} {SIGIR}
  Conference on Research and Development in Information Retrieval, Shinjuku,
  Tokyo, Japan, August 7-11, 2017}, N.~Kando, T.~Sakai, H.~Joho, H.~Li, A.~P.
  de~Vries, and R.~W. White, Eds.\hskip 1em plus 0.5em minus 0.4em\relax {ACM},
  2017, pp. 355--364. [Online]. Available:
  \url{https://doi.org/10.1145/3077136.3080777}
\BIBentrySTDinterwordspacing

\bibitem{paper2cite3}
\BIBentryALTinterwordspacing
P.~Covington, J.~Adams, and E.~Sargin, ``Deep neural networks for youtube
  recommendations,'' in \emph{Proceedings of the 10th {ACM} Conference on
  Recommender Systems, Boston, MA, USA, September 15-19, 2016}, S.~Sen,
  W.~Geyer, J.~Freyne, and P.~Castells, Eds.\hskip 1em plus 0.5em minus
  0.4em\relax {ACM}, 2016, pp. 191--198. [Online]. Available:
  \url{https://doi.org/10.1145/2959100.2959190}
\BIBentrySTDinterwordspacing

\bibitem{paper1cite16}
\BIBentryALTinterwordspacing
Y.~Juan, Y.~Zhuang, W.~Chin, and C.~Lin, ``Field-aware factorization machines
  for {CTR} prediction,'' in \emph{Proceedings of the 10th {ACM} Conference on
  Recommender Systems, Boston, MA, USA, September 15-19, 2016}, S.~Sen,
  W.~Geyer, J.~Freyne, and P.~Castells, Eds.\hskip 1em plus 0.5em minus
  0.4em\relax {ACM}, 2016, pp. 43--50. [Online]. Available:
  \url{https://doi.org/10.1145/2959100.2959134}
\BIBentrySTDinterwordspacing

\bibitem{paper1cite7}
\BIBentryALTinterwordspacing
C.~Cheng, F.~Xia, T.~Zhang, I.~King, and M.~R. Lyu, ``Gradient boosting
  factorization machines,'' in \emph{Eighth {ACM} Conference on Recommender
  Systems, RecSys '14, Foster City, Silicon Valley, CA, {USA} - October 06 -
  10, 2014}, A.~Kobsa, M.~X. Zhou, M.~Ester, and Y.~Koren, Eds.\hskip 1em plus
  0.5em minus 0.4em\relax {ACM}, 2014, pp. 265--272. [Online]. Available:
  \url{https://doi.org/10.1145/2645710.2645730}
\BIBentrySTDinterwordspacing

\bibitem{paper1cite40}
\BIBentryALTinterwordspacing
J.~Xiao, H.~Ye, X.~He, H.~Zhang, F.~Wu, and T.~Chua, ``Attentional
  factorization machines: Learning the weight of feature interactions via
  attention networks,'' in \emph{Proceedings of the Twenty-Sixth International
  Joint Conference on Artificial Intelligence, {IJCAI} 2017, Melbourne,
  Australia, August 19-25, 2017}, C.~Sierra, Ed.\hskip 1em plus 0.5em minus
  0.4em\relax ijcai.org, 2017, pp. 3119--3125. [Online]. Available:
  \url{https://doi.org/10.24963/ijcai.2017/435}
\BIBentrySTDinterwordspacing

\bibitem{paper2cite5}
\BIBentryALTinterwordspacing
Y.~Qu, H.~Cai, K.~Ren, W.~Zhang, Y.~Yu, Y.~Wen, and J.~Wang, ``Product-based
  neural networks for user response prediction,'' in \emph{{IEEE} 16th
  International Conference on Data Mining, {ICDM} 2016, December 12-15, 2016,
  Barcelona, Spain}, F.~Bonchi, J.~Domingo{-}Ferrer, R.~Baeza{-}Yates, Z.~Zhou,
  and X.~Wu, Eds.\hskip 1em plus 0.5em minus 0.4em\relax {IEEE} Computer
  Society, 2016, pp. 1149--1154. [Online]. Available:
  \url{https://doi.org/10.1109/ICDM.2016.0151}
\BIBentrySTDinterwordspacing

\bibitem{paper1cite38}
\BIBentryALTinterwordspacing
R.~Wang, B.~Fu, G.~Fu, and M.~Wang, ``Deep {\&} cross network for ad click
  predictions,'' in \emph{Proceedings of the ADKDD'17, Halifax, NS, Canada,
  August 13 - 17, 2017}.\hskip 1em plus 0.5em minus 0.4em\relax {ACM}, 2017,
  pp. 12:1--12:7. [Online]. Available:
  \url{https://doi.org/10.1145/3124749.3124754}
\BIBentrySTDinterwordspacing

\bibitem{paper1cite19}
\BIBentryALTinterwordspacing
J.~Lian, X.~Zhou, F.~Zhang, Z.~Chen, X.~Xie, and G.~Sun, ``xdeepfm: Combining
  explicit and implicit feature interactions for recommender systems,'' in
  \emph{Proceedings of the 24th {ACM} {SIGKDD} International Conference on
  Knowledge Discovery {\&} Data Mining, {KDD} 2018, London, UK, August 19-23,
  2018}, Y.~Guo and F.~Farooq, Eds.\hskip 1em plus 0.5em minus 0.4em\relax
  {ACM}, 2018, pp. 1754--1763. [Online]. Available:
  \url{https://doi.org/10.1145/3219819.3220023}
\BIBentrySTDinterwordspacing

\bibitem{paper1cite2}
\BIBentryALTinterwordspacing
D.~Bahdanau, K.~Cho, and Y.~Bengio, ``Neural machine translation by jointly
  learning to align and translate,'' in \emph{3rd International Conference on
  Learning Representations, {ICLR} 2015, San Diego, CA, USA, May 7-9, 2015,
  Conference Track Proceedings}, Y.~Bengio and Y.~LeCun, Eds., 2015. [Online].
  Available: \url{http://arxiv.org/abs/1409.0473}
\BIBentrySTDinterwordspacing

\bibitem{paper2cite25}
\BIBentryALTinterwordspacing
S.~Zhai, K.~Chang, R.~Zhang, and Z.~M. Zhang, ``Deepintent: Learning attentions
  for online advertising with recurrent neural networks,'' in \emph{Proceedings
  of the 22nd {ACM} {SIGKDD} International Conference on Knowledge Discovery
  and Data Mining, San Francisco, CA, USA, August 13-17, 2016},
  B.~Krishnapuram, M.~Shah, A.~J. Smola, C.~C. Aggarwal, D.~Shen, and
  R.~Rastogi, Eds.\hskip 1em plus 0.5em minus 0.4em\relax {ACM}, 2016, pp.
  1295--1304. [Online]. Available:
  \url{https://doi.org/10.1145/2939672.2939759}
\BIBentrySTDinterwordspacing

\bibitem{dest_1}
Y.~Lassoued, J.~Monteil, Y.~Gu, G.~Russo, R.~Shorten, and M.~Mevissen, ``A
  hidden markov model for route and destination prediction,'' in \emph{20th
  {IEEE} International Conference on Intelligent Transportation Systems, {ITSC}
  2017, Yokohama, Japan, October 16-19, 2017}.\hskip 1em plus 0.5em minus
  0.4em\relax {IEEE}, 2017, pp. 1--6.

\bibitem{dest_2}
A.~V. Khezerlou, X.~Zhou, L.~Tong, Y.~Li, and J.~Luo, ``Forecasting gathering
  events through trajectory destination prediction: {A} dynamic hybrid model,''
  \emph{{IEEE} Trans. Knowl. Data Eng.}, vol.~33, no.~3, pp. 991--1004, 2021.

\bibitem{dest_3}
X.~Zhang, Z.~Zhao, Y.~Zheng, and J.~Li, ``Prediction of taxi destinations using
  a novel data embedding method and ensemble learning,'' \emph{{IEEE} Trans.
  Intell. Transp. Syst.}, vol.~21, no.~1, pp. 68--78, 2020.

\bibitem{dest_4}
W.~Wang, J.~Chen, J.~Wang, J.~Chen, J.~Liu, and Z.~Gong, ``Trust-enhanced
  collaborative filtering for personalized point of interests recommendation,''
  \emph{{IEEE} Trans. Ind. Informatics}, vol.~16, no.~9, pp. 6124--6132, 2020.

\end{thebibliography}

\begin{IEEEbiography}[{\includegraphics[width=1in,height=1.25in,clip,keepaspectratio]{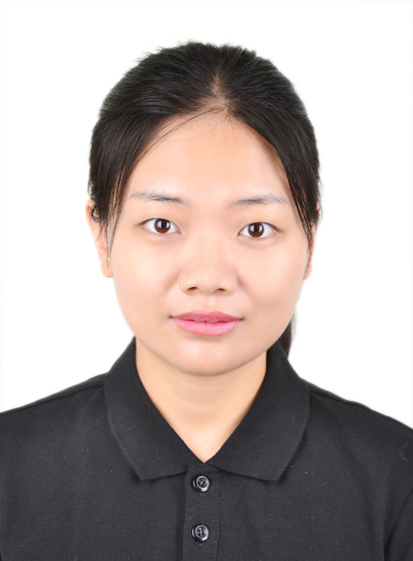}}]{Yu LI} 
received the Ph.D. degree from the Hong Kong Polytechnic University in 2015. Her research focuses on crowdsourcing, spatial recommendation, database optimization and cloud computing. 
She works in Hangzhou Dianzi University in China currently.
Before that, she worked as a postdoc researcher in the Hong Kong Polytechnic University.
\protect
\end{IEEEbiography}

\begin{IEEEbiography}[{\includegraphics[width=1in,height=1.25in,clip,keepaspectratio]{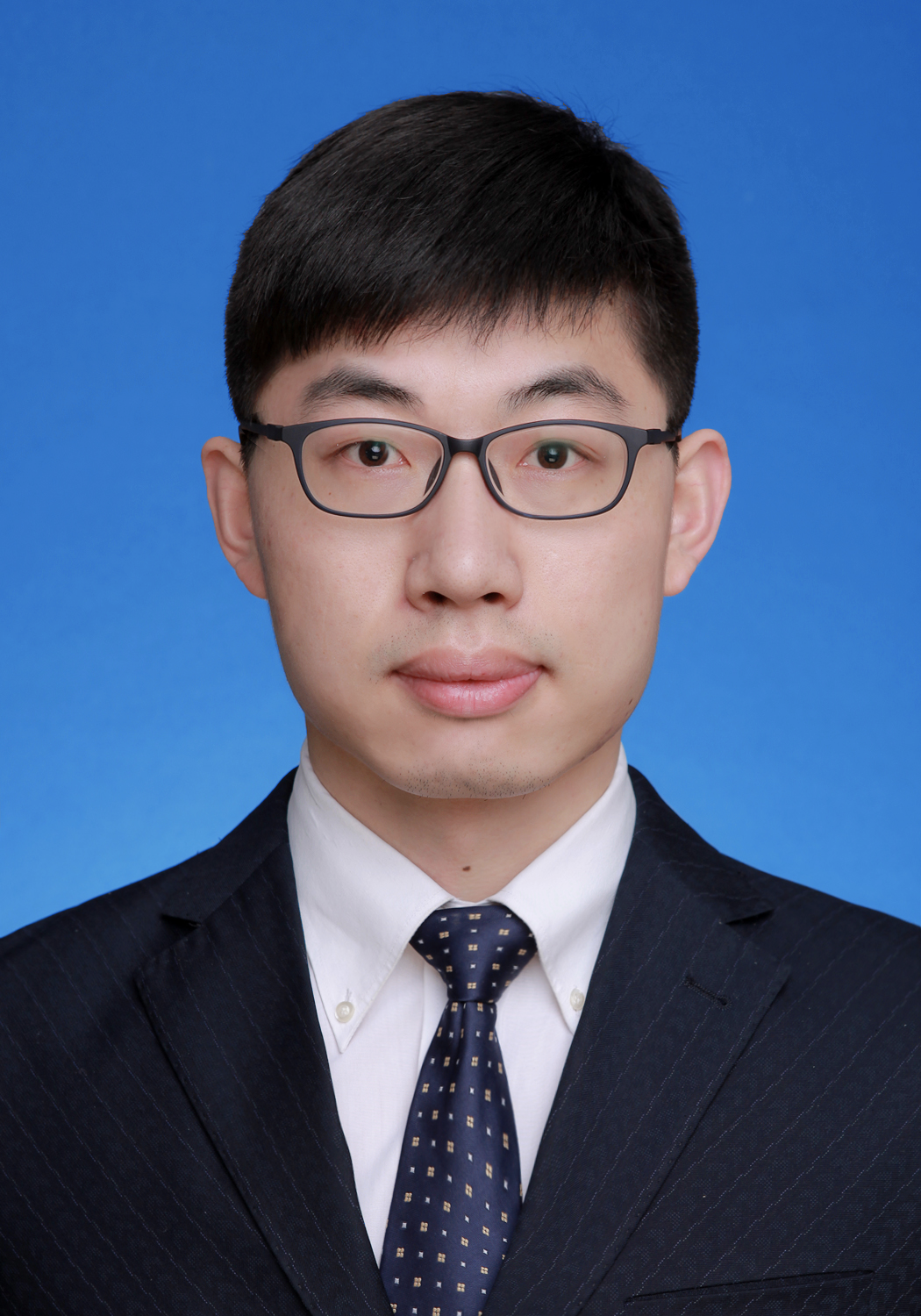}}]{Fei Xiong} 
received the B.E. degree in automation major from the Wuhan University, Wuhan, China, in 2016, and the M.E. degree in control engineering from the Shanghai Jiaotong University, Shanghai, China, in 2019. He is currently working in Alibaba group as a senior algorithm engineer, participating in the development of recommendation system and search engine. His research interests include data mining, machine learning, and applications in artificial intelligence.
\protect
\end{IEEEbiography}

\begin{IEEEbiography}[{\includegraphics[width=1in,height=1.25in,clip,keepaspectratio]{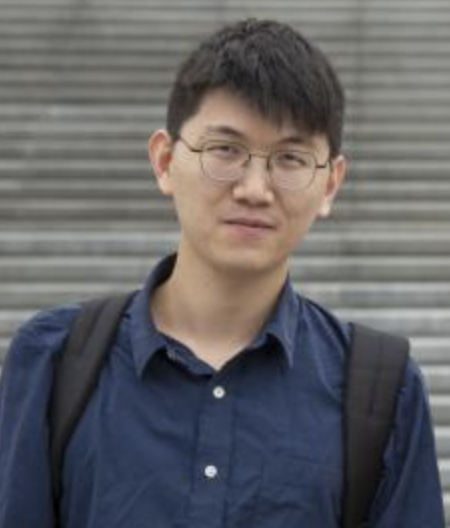}}]{Ziyi Wang} 
received the M.E. degree in Computer Science and Technology from Nanjing University in June 2019. He is currently working in Alibaba, and his current research interests mainly include personalized ranking algorithm in eCommerce (travel scenario) search and recommendation.
\protect
\end{IEEEbiography}

\begin{IEEEbiography}[{\includegraphics[width=1in,height=1.25in,clip,keepaspectratio]{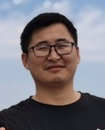}}]{Zulong Cheng} 
received the M.E. degree in Information Engineering from Northeastern University. He worked
in Baidu in 2013. After that, he joined Alibaba
in 2014, working on personalized click-through rate estimation and crowd mining algorithms.
He is currently working in Fliggy, focusing on
search recommendation and advertising.
He has published papers in KDD, CIKM, and WWW.
\protect
\end{IEEEbiography}

\begin{IEEEbiography}[{\includegraphics[width=1in,height=1.25in,clip,keepaspectratio]{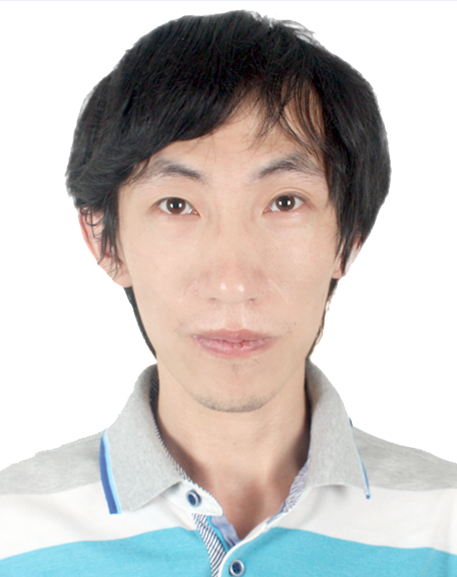}}]{Chuanfei Xu} 
received the BE degree from the Shenyang University of Technology in 2007 and the ME and Phd degrees in computer science from Northeastern University in 2009 and 2013. 
He is currently a postdoctoral research fellow in Concordia University, Canada. 
His research interests include data mining, NLP and uncertain data management.
\protect
\end{IEEEbiography}

\begin{IEEEbiography}[{\includegraphics[width=1in,height=1.25in,clip,keepaspectratio]{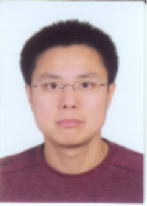}}]{Yuyu Yin} 
received the Ph.D. degree in computer science from Zhejiang University in 2010. He is currently a Professor with the College of Computer, Hangzhou Dianzi University. He is also a Supervisor of master’s students with the School of Computer Engineering and Science, Shanghai University, Shanghai, China. He has published more than 40 articles in journals and refereed conferences, such as Sensors, Entropy, International Journal of Software Engineering and Knowledge Engineering, Mobile Information Systems, ICWS, and SEKE. His research interests include service computing, cloud computing, and business process management. He has organized more than ten international conferences and workshops, such as FMSC 2011/2017 and DISA 2012 and 2017/2018. He has served as a Guest Editor for the Journal of Information Science and Engineering and International Journal of Software Engineering and Knowledge Engineering and a reviewer for the IEEE Transactions on Industry Informatics, Journal of Database Management, and Future Generation Computer Systems. He is also a member of the China Computer Federation (CCF) and the CCF Service Computing Technical Committee.
\protect
\end{IEEEbiography}

\begin{IEEEbiography}[{\includegraphics[width=1in,height=1.25in,clip,keepaspectratio]{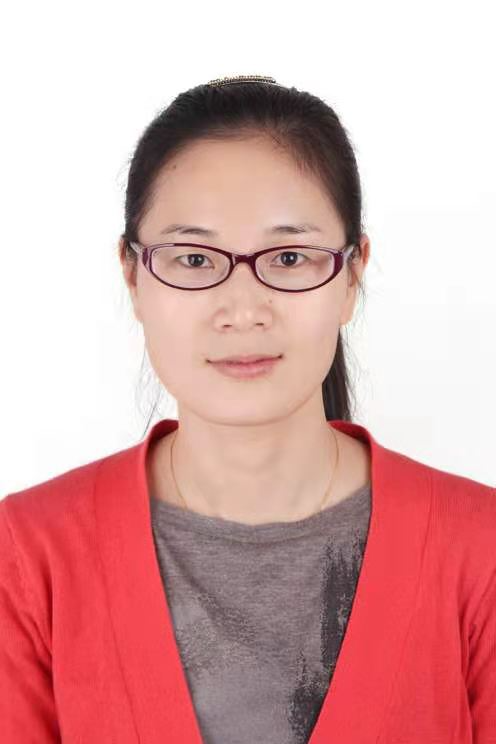}}]{Li Zhou} 
received the master degree from Hangzhou Dianzi 
University. She is currently working as 
an associate professor in Hangzhou Dianzi 
University. She has hosted the Natural Science
Foundation of Zhejiang Province, the sub-projects
of the National Natural Fund and the sub-projects
of Zhejiang Province's key R\&D projects. She has
published over 30 papers indexed by SCI/EI.

\protect
\end{IEEEbiography}

\end{document}